\documentclass[10pt,twocolumn,letterpaper]{article}

\usepackage[pagenumbers]{arxivpaper}

\renewcommand{\paragraph}[1]{\vspace{.5em}\noindent\textbf{#1}}

\newcommand{\releasenote}{}

\newcommand{\gtcatalog}{The list of mechanics and the logging code build on a concurrent study of the same games~\cite{concurrent2026perception}, which uses them to judge the text reports of language-model agents.
We add the cells that each mechanic changes and the exposure index, and we score the next-frame predictions of a world model on human recordings.}

\usepackage{xspace}
\newcommand{\method}{WARC\xspace}
\newcommand{\std}[1]{\,{\scriptsize$\pm$#1}}   %
\newcommand{\evalsuite}{MechEval\xspace}

\AtEndPreamble{%
  \crefname{appendix}{Appendix}{Appendices}%
  \Crefname{appendix}{Appendix}{Appendices}%
  \Crefname{figure}{Fig.}{Figs.}%
  \Crefname{table}{Tab.}{Tabs.}%
}
\usepackage[above]{placeins}   %
\usepackage{capt-of}           %
\usepackage{colortbl}           %

\usepackage{algorithm}
\usepackage{algpseudocode}

\DeclareRobustCommand{\chg}[1]{#1}
\usepackage{tcolorbox}
\tcbuselibrary{breakable}
\definecolor{linkgreen}{HTML}{6E9A2E}
\usepackage[pagebackref,breaklinks,colorlinks,allcolors=linkgreen]{hyperref}

\renewcommand{\releasenote}{Code is available at \url{https://github.com/third-intelligence/WARC}.}

\newcommand{\suppcopynote}{}

\title{How Many Exposures Does a World Model Need?\\Learning Mechanics in Unseen Games at Test Time}

\author{Ku Onoda$^{1,2}$ \quad Haruto Kawano$^{1,2}$ \quad Shota Takashiro$^{1,2}$ \quad Kohei Hayashi$^{1,2}$\\
$^1$Third Intelligence, Inc. \quad $^2$The University of Tokyo
}

\begin{document}
\twocolumn[{%
\renewcommand\twocolumn[1][]{#1}%
\maketitle
\vspace{-1.0em}
\begin{center}
  \includegraphics[width=\textwidth]{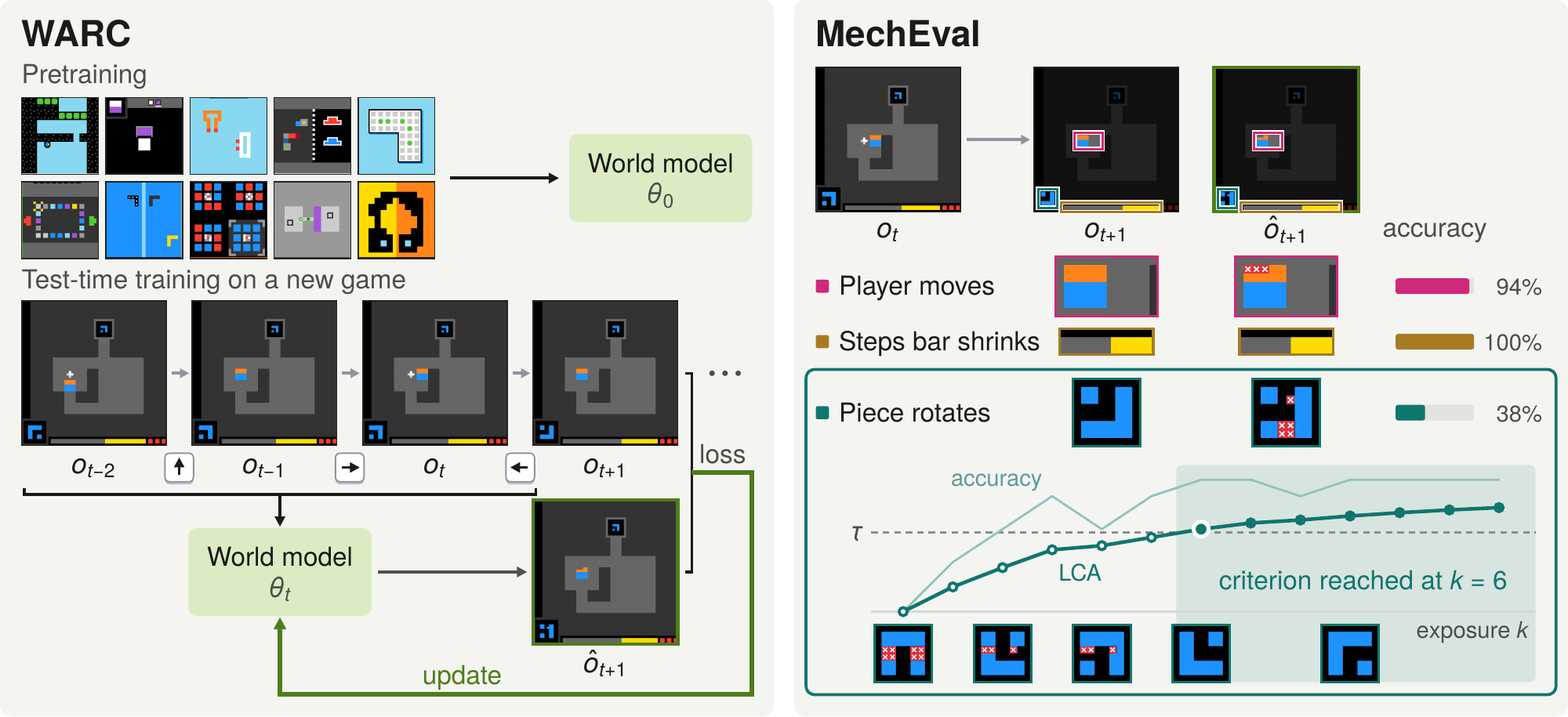}
  \captionof{figure}{\textbf{\method and \evalsuite.} Left: \method is pretrained on ARC-AGI-3 games.
  In a new game, it predicts each next frame from recent frames and actions, then updates its weights.
  Right: \evalsuite reads from the game program which cells each mechanic changes, and scores the predictions on these cells at each exposure $k$.
  The learning curve area LCA$_m$@$k$ (bold) is the mean of the accuracy (faint) up to $k$.
  A mechanic reaches the criterion once LCA$_m$@$k$ stays at or above a threshold $\tau$.
  The curve is illustrative.
  }
  \label{fig:method}
\end{center}
\vspace{0.6em}
}]
\begin{abstract}
Learning the rules, or mechanics, of unseen environments from limited interaction is a central challenge for neural world models.
Yet task success also depends on goal discovery and exploration, while aggregate prediction accuracy can hide failures on infrequently encountered mechanics.
We introduce \evalsuite{} to measure how much experience a world model needs to predict each mechanic's effects.
Using annotations from game programs, \evalsuite{} tracks 69 recurring mechanics across 20 public ARC-AGI-3 games and measures accuracy on the cells each mechanic changes as a function of its prior occurrences, or exposures.
Experiments with \method{} (World model for ARC-AGI-3), a vision transformer pretrained on other games, show that test-time adaptation substantially improves prediction in unseen games.
On learning curves averaged across recordings, \method brings more than twice as many mechanics as any baseline to a criterion of 70\% accuracy, averaged over the exposures so far, within 100 exposures.
The adapted model also improves action efficiency: \method-Agent, which pairs \method{} with a fixed graph explorer, triples the explorer's official efficiency score on public ARC-AGI-3 games, with most of the gain coming from test-time training.
\releasenote

\end{abstract}    
\section{Introduction}

Adapting to environments with unseen dynamics remains a major challenge for neural world models.
World models predict the consequences of actions to support planning and learning from imagined experience~\cite{ha2018worldmodels,hafner2023dreamerv3}.
However, the dynamics encountered at test time may differ substantially from those seen during training.
Previous studies have demonstrated adaptation to changes in physical dynamics, including variations in terrain slope and mass~\cite{nagabandi2018adapt,wang2026adajepa}.
\citet{gao2025adaworld} have addressed unseen controls by reusing pretrained latent action representations.
Despite these advances, how quickly neural world models learn to
predict individual rules in unseen environments remains open.

Interactive games provide a controlled setting for studying this challenge.
Their dynamics are governed by a set of rules known as \emph{mechanics}.
Each mechanic specifies the conditions under which an effect occurs and the change it produces.
For example, a tile may rotate any object placed on it regardless of where the tile appears.
Learning such a mechanic requires identifying its conditions and effects to predict what happens in new situations~\cite{kansky2017schema,pouncy2021model,tsividis2021theory}.
However, different mechanics may be encountered at very different frequencies.
A model may observe movement many times while encountering a particular rotation only a few times.
Understanding adaptation therefore requires measuring how much experience a model needs to learn each rule.
This motivates our central question.
\textbf{How many exposures to a mechanic in an unseen game does a world model need to predict its effects?}
We define an \emph{exposure} as a single occurrence of a mechanic.

Answering this question requires measuring prediction accuracy for each mechanic in isolation.
In an unseen game, task success depends not only on learning the dynamics but also on discovering the goal and exploring the environment~\cite{ying2025assessing,lambert2020objective}.
Action selection also affects the outcome.
Task performance alone therefore cannot isolate the world model.
Next-frame prediction accuracy provides a more direct measure of dynamics learning, but aggregate accuracy entangles performance across different mechanics.
Whole-frame accuracy is often dominated by unchanged cells.
Even accuracy on changed cells can be dominated by frequently occurring effects such as movement.

To address this gap, we study ARC-AGI-3~\cite{arcprize2026techreport}, a benchmark of interactive games rendered as $64\times64$ grids of colored cells.
Each game is designed to be novel, and its rules and goals must be discovered through play.
We introduce \textbf{\evalsuite{}}, an evaluation protocol that measures prediction accuracy for individual mechanics as a function of prior exposures.
Using annotations derived from game programs, \evalsuite{} tracks 69 recurring mechanics across 20 public games.
The annotations identify which mechanics fire in each transition and which cells they change.
We measure per-mechanic prediction accuracy according to its number of prior exposures.
To investigate how neural world models learn mechanics in unseen games, we develop \textbf{WARC (World model for ARC-AGI-3)}, a 35M-parameter vision transformer that extends a visual prediction and test-time training approach developed for static ARC-AGI benchmarks~\cite{hu2025varc,sun2020ttt} to sequential game interactions.
This simple design enables controlled comparisons of pretraining and update timing as well as which parameters are updated.

Our results show that test-time adaptation substantially improves the prediction of mechanics in unseen games.
On learning curves averaged across recordings, \method brings more than twice as many mechanics as any baseline to a criterion of 70\% accuracy, averaged over the exposures so far, within 100 exposures.
Updating after every transition improves accuracy at early exposures even when the average number of gradient steps per transition is held fixed.
Finally, we integrate WARC into a fixed graph explorer.
This substantially improves the action efficiency score on 20 public games.
Our contributions are summarized below.
\begin{itemize}
    \item \textbf{An evaluation of learning by mechanic and exposure.}
    \evalsuite{} combines annotations from game programs with predictions to track the learning of individual mechanics under shared experience.

    \item \textbf{A controlled study of predictive adaptation.}
    We quantify how quickly mechanics in unseen games become predictable.
    We also investigate how pretraining, the adaptation mechanism, and the choice of parameters to update affect learning.

    \item \textbf{A downstream test with a fixed explorer.}
    We demonstrate that adapting WARC to unseen game dynamics improves the action efficiency of a graph explorer.
\end{itemize}

\section{Related Work}
\label{sec:related}

\paragraph{Adapting a world model to a new environment.}
A world model is usually learned from the environment it models, as in model-based agents such as DreamerV3~\cite{hafner2023dreamerv3,micheli2023iris,zhang2023storm,heins2025axiom}.
Pretrained world models transfer across games and tasks but are still finetuned on thousands of transitions or more of each new one~\cite{xu2023xtra,cheng2025jowa,hansen2025newt,hansen2026hallucination}. AdaWorld adapts from few interactions by pretraining with latent actions~\cite{gao2025adaworld}.
In-context methods infer a context from recent transitions~\cite{nagabandi2018adapt,lee2020cadm,prasanna2024crssm,roder2025dali}, amortize the update into a hypernetwork or a mixture of experts~\cite{palafox2026claw,jang2026tmow}, or learn from a synthetic prior~\cite{ferreira2024oswm}.
Gradient-based methods update the model at test time. AdaJEPA does so inside the planning loop~\cite{wang2026adajepa}, JEPA-TTT across test episodes~\cite{zhang2026jepattt}, others on explored transitions or in residual modules~\cite{wan2026worldagen,soni2026sandwich}, and CAVIA only on a context vector~\cite{zintgraf2019cavia}.
We compare these design choices under a common protocol on ARC-AGI-3 and study which of them make a world model learn fast.

\paragraph{Learning the rules of a game.}
Game description languages such as VGDL~\cite{schaul2013vgdl} and Autumn~\cite{das2023autumn} write a game as a set of rules.
Theory-based agents infer such a rule set from play as a program~\cite{tsividis2021theory}, and language-model agents, including several published agents on ARC-AGI-3, write and repair it as code~\cite{tang2024worldcoder,rodionov2026ewm,skoutnev2026twin}.
These methods recover the rules explicitly.
Our mechanics are such rules, which \evalsuite reads from the game programs, and we ask how many exposures a neural world model needs to predict their effects.

\paragraph{Evaluating world models.}
Evaluation of world models has been extended from visual fidelity to physical understanding~\cite{motamed2025physicsiq} and the consistency of game mechanics~\cite{lin2026stateplay}.
ScratchWorld credits a prediction only when it computes the fields that an action changes~\cite{lin2026scratchworld}, and AutumnBench evaluates prediction and planning after a reward-free interaction phase~\cite{warrier2025autumnbench}.
\citet{ying2025assessing} propose new games under a trial budget to assess how efficiently a world model adapts.
In contrast, \evalsuite scores each mechanic at each exposure while the model learns, with ground-truth labels from the game programs.
\Cref{app:related} covers other work.

\section{Preliminaries}
\label{sec:setting}

\paragraph{Games.}
Each ARC-AGI-3 game is a turn-based environment governed by a deterministic program~\cite{arcprize2026techreport}.
At every turn, the agent observes a frame $o_t$ and takes an action $a_t$.
A frame $o_t$ is a $64{\times}64$ grid of cells in 16 colors, and an action is one of up to five keys, a click on a cell, or in some games an undo.
The program keeps a hidden state that changes only when the agent acts, $s_{t+1} = T(s_t, a_t)$, where $s_t$ is the state and $T$ the game's transition function.
The frame is rendered from the state, $o_t = \rho(s_t)$.
Each game consists of a sequence of levels.
A level ends once its hidden goal is reached, and the next level applies the same dynamics to a new layout.
Of its games, 25 are public with their programs, and the others are kept private for evaluation.
The setting is discrete and deterministic, and since the frame shows the whole board, it is nearly fully observable.

\paragraph{Mechanics.}
As in game description languages~\cite{schaul2013vgdl,das2023autumn,tsividis2021theory}, a \emph{mechanic} $m$ pairs a binary condition $c_m$ on the state and action with a change $e_m$ to the objects it acts on.
It fires whenever $c_m=1$.
The transition function is then the combined effect of the mechanics that fire,
\begin{equation}
  T(s_t, a_t) = s_t \oplus \bigcup_{m \in M_t} e_m(s_t, a_t),
  \label{eq:mechanics}
\end{equation}
where $M_t=\{m : c_m(s_t,a_t)=1\}$ is the set of mechanics that fire and $\oplus$ applies their changes to the state in the order the program defines.
Learning a game's dynamics therefore means learning its mechanics.
Mechanics also recur during play, at other positions and in later levels.

Reaching the goal is also a mechanic.
Its condition is the goal state, and its effect ends the level.
It fires only once per level, too rarely to measure how it is learned, so we leave it out of the evaluation.

\section{\texorpdfstring{\evalsuite}{MechEval}}
\label{sec:mecheval}

Our question is how many exposures a model needs before it predicts the effect of a mechanic in a new game.
Accuracy over all cells of the frame cannot answer this question for two reasons.
First, copying the last frame already predicts 98.8\% of the cells correctly, because most cells do not change.
Second, an average over a game is dominated by its most frequent mechanics.
We therefore introduce \evalsuite, which consists of an annotator and a scorer.
The annotator takes a trajectory $\{(o_t,a_t)\}$ and the identity of its game, and returns, for every step $t$, the mechanics $M_t$ that fire and the set $C_m(t)$ of cells that each one changes,
\begin{equation}
  \mathcal{A}_t = \big\{(m,\,C_m(t)) : m\in M_t\big\}.
  \label{eq:annotator}
\end{equation}
The scorer takes $\mathcal{A}_t$ for every step and a model's predictions, and measures the accuracy on each mechanic's changed cells as a function of its prior exposures (\cref{fig:method}).
It uses nothing else from the environment, so models can be evaluated in the same way on any simulator that reports $\mathcal{A}_t$.

For ARC-AGI-3, we build the annotator by reverse engineering the public game programs.
For each mechanic $m$, we find the code that implements its change $e_m$ in \cref{eq:mechanics} and log when it runs and which objects it changes.
We then run the program again on the actions of a trajectory, which gives the mechanics $M_t$ that fire at each step.
The cells of their objects that change give each $C_m(t)$, and the two form $\mathcal{A}_t$.
\gtcatalog{}
\Cref{app:mecheval} gives the details.

\paragraph{Benchmark.}
\evalsuite covers 20 of the 25 public games, with 69 mechanics.
We annotate 279 human recordings~\cite{arcprize2026humandata}, each truncated when level 2 is first completed.
A change that happens only once or twice, such as picking up the only key of a level or reaching its goal, leaves no later exposure to predict.
We therefore keep a mechanic only if it fires at least three times in the median recording that contains it.

\paragraph{Protocol.}
The protocol evaluates a world model without a policy.
It runs on recorded trajectories, where the actions are those of the human player and the model selects none.
At every step, the model predicts the next frame $\hat{o}_{t+1}$ before it learns from that transition~\cite{dawid1984prequential}.
Every model therefore gets the same exposures in the same order.

\paragraph{Accuracy by exposure.}
We index the exposures to a mechanic from 0, so the index of mechanic $m$ at step $t$ is the number of earlier steps at which it fired, including steps whose change is not visible,
\begin{equation}
  k_m(t) = \sum_{i<t} \mathbf{1}\big[m\in M_i\big].
  \label{eq:exposure}
\end{equation}
An exposure to $m$ at step $t$ whose change is visible, $C_m(t)\neq\emptyset$, is scored by the share of its changed cells that the model predicts correctly,
\begin{equation}
  \frac{1}{|C_m(t)|}\sum_{g\in C_m(t)} \mathbf{1}\big[\hat{o}_{t+1}(g)=o_{t+1}(g)\big].
  \label{eq:acc}
\end{equation}
The average over a mechanic's exposures with the same $k$ gives its \emph{learning curve}.
In \cref{fig:method} (right), at step $t$, predictions are nearly perfect for the player and steps bar but only partly correct for piece rotation.

\paragraph{Learning curve area.}
We summarize each learning curve using its learning curve area (LCA), a metric used in continual learning to measure how quickly a learner acquires a task~\cite{chaudhry2019agem}.
For a mechanic $m$, LCA$_m$@$k$ is the mean of its learning curve over exposures 0 through $k$ at which the change is visible, and it keeps its final value beyond the last exposure.
A model that becomes accurate after fewer exposures therefore has a higher LCA$_m$@$k$ than one that reaches the same accuracy later.
We report LCA@$k$ by averaging LCA$_m$@$k$ across all mechanics.

\paragraph{Criterion.}
A mechanic $m$ reaches the criterion at the earliest exposure $k$ from which LCA$_m$@$k$ remains at or above a threshold $\tau$ through its final exposure.
Mechanics whose LCA$_m$@$k$ at the final exposure is below $\tau$ are marked as not reached (n.r.).
We report Reached@$k$ as the cumulative share of all mechanics that have reached the criterion by exposure $k$.
We set $\tau=70\%$, and \cref{app:curves} varies it.

\section{\texorpdfstring{\method}{WARC}}
\label{sec:method}

To study what makes a world model learn fast, we need to isolate the effects of individual design choices.
World models such as DreamerV3~\cite{hafner2023dreamerv3} couple an encoder, a recurrent latent state, and a decoder through several losses, making prediction errors difficult to attribute to individual components.
We therefore build on VARC~\cite{hu2025varc}, which trains a vision transformer with a single pixel-space cross-entropy loss and performs well on ARC-AGI-1 and ARC-AGI-2.
These static puzzles use grids of colored cells, as do the game frames of ARC-AGI-3.
After reviewing VARC (\cref{sec:varc}), we describe how \method adapts it to games through action-conditioned next-frame prediction (\cref{sec:model}) and online test-time training (\cref{sec:ttt}).

\subsection{Background: VARC}
\label{sec:varc}

VARC predicts the color of each cell in an output grid from an input grid, conditioned on a learned task token for the puzzle.
For each new puzzle, test-time training~\cite{sun2020ttt,akyurek2024ttt} updates the network and a new task token on the two to four demonstrations and their rotated, flipped, and recolored copies before predicting the test output.

\subsection{Model}
\label{sec:model}

The first change makes \method a world model, which predicts the next frame of a game instead of the output grid of a puzzle.
It is a vision transformer~\cite{dosovitskiy2021vit} with parameters $\theta$ that maps the history $x_t=(o_{t-h+1:t},\,a_{t-h+1:t})$ of the last $h$ frames and actions to a distribution $p_\theta(o_{t+1}\mid x_t)$ over the next frame.
Both pretraining and test-time training minimize the negative log-likelihood of the observed next frame, which is the sum of the cross-entropies of its cells,
\begin{equation}
  \mathcal{L}_t(\theta) = -\log p_\theta(o_{t+1}\mid x_t).
  \label{eq:loss}
\end{equation}

The second change is in the architecture.
(i) \method uses $3{\times}2{\times}2$ spatiotemporal patches to jointly embed $h=3$ frames as $32{\times}32$ tokens.
(ii) We introduce spatially aligned action conditioning.
The current action type and click coordinates condition every transformer block through adaLN~\cite{peebles2023dit}.
We align clicks with the observation grid by adding learned vectors to the clicked patch token, identifying the target cell within its $2{\times}2$ patch.
(iii) \method requires no task-specific learned token, because removing it did not lower accuracy in preliminary experiments.

\subsection{Learning on the Stream}
\label{sec:ttt}

The third and main change is in how \method learns at test time.
In VARC, the data are fixed. The demonstrations of a puzzle are given at once and there is a single test output, so the model trains on them as a batch before it predicts.
In \method, the data arrive as a stream. Each transition is a new prediction target, so \method keeps learning online while it predicts, as in online test-time training~\cite{sun2020ttt,wang2022cotta,wang2023tttstreams}.
It predicts transition $t$ with its current parameters $\theta_t$ before training on it.
It then takes several gradient steps on the loss of \cref{eq:loss}, each on a new minibatch sampled from a replay buffer of the transitions observed so far.

Because the update follows every transition, \method has already trained on all $k$ earlier exposures of a mechanic when it predicts exposure $k$.
Each minibatch mixes the most recent transitions with older ones sampled at random, to reduce forgetting.
All parameters are updated, so a new mechanic can change any part of the model.
Each transition is augmented with its three rotations.
\Cref{app:model} gives the architecture and hyperparameters.

The input of \method covers only the last $h$ frames and actions, so a transition outside this history affects a prediction only through the updates it has caused.
The weights $\theta_t$ thus play the role of a recurrent state, as the weights of the inner model do in test-time training layers~\cite{sun2025ttt}.

\section{Experiments on \texorpdfstring{\evalsuite}{MechEval}}
\label{sec:experiments}

\begin{figure*}[t]
  \centering
  \includegraphics[width=\linewidth]{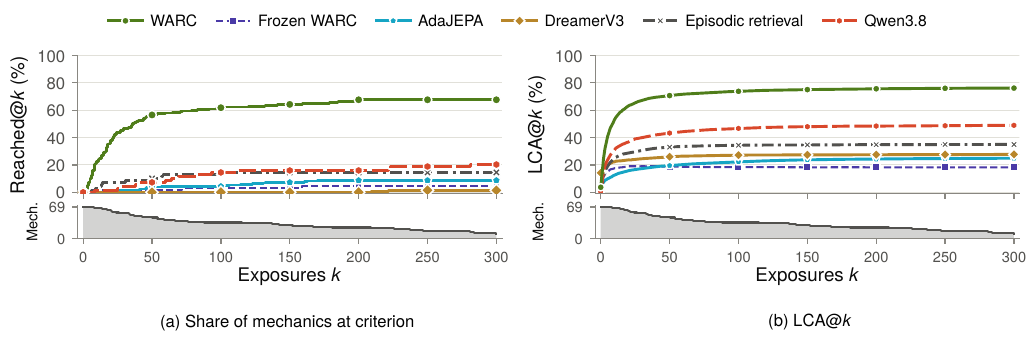}
  \caption{\textbf{Learning to predict the mechanics of unseen games.}
  (a) Reached@$k$: cumulative share of all 69 mechanics that have reached the criterion by exposure $k$ (LCA$_m$@$k$ remains at or above 70\% through the final exposure).
  The strips count mechanics whose recordings reach exposure $k$.
  (b) LCA@$k$: learning curve area averaged across all mechanics.}
  \label{fig:cumulative}
\end{figure*}

\subsection{Setup}
\label{sec:setup}

\paragraph{Protocol.}
For all models pretrained in our experiments, pretraining and evaluation use disjoint games.
We split the 25 public games into three folds and evaluate each fold using models pretrained on human recordings from the other two (\cref{app:model}).
Test-time training proceeds independently for each recording: the model starts from its pretrained checkpoint, learns as it progresses through the recording, and is discarded afterward.
For methods updated at test time, we average results over three test-time seeds.
In addition to the metrics in \cref{sec:mecheval}, we report the \emph{false-edit rate} (FE), the fraction of unchanged cells predicted to change, including on steps where a move is blocked (\cref{app:novelty}).

\paragraph{Baselines.}
We compare \method with four baselines, whose implementation details and further results are in \cref{app:baselines}.
All methods are evaluated on the same observation–action trajectories and scored on the same next-frame targets.
\method, AdaJEPA, and DreamerV3 take 20 gradient steps on minibatches of eight after every transition.
The rest of each test-time recipe was selected on the five public games not covered by \evalsuite, so that no evaluated game influenced the choice (\cref{app:model}).
\begin{itemize}
  \item \textbf{AdaJEPA}~\cite{wang2026adajepa} is a latent world model. We match its inputs and outputs to \method and add an inverse-dynamics loss~\cite{sobal2025pldm}, without which it predicted few changed cells.
  \item \textbf{DreamerV3}~\cite{hafner2023dreamerv3} is a world model whose recurrent state is recomputed over the last 64 frames.
  \item \textbf{Episodic retrieval} stores the transitions observed so far and predicts each cell that followed the same $3{\times}3$ neighborhood under the same action type in an earlier transition. It tests whether the mechanics can be looked up rather than learned.
  \item \textbf{LLM (Qwen3.8)} uses Qwen3.8-Flash~\cite{qwen2026flashnext}, a general-purpose pretrained language model with a context of one million tokens. We chose it for this long context. It predicts the next frame in context from the frames and actions so far, given as text, without parameter updates.
\end{itemize}
AdaJEPA and DreamerV3 are pretrained on the same folds as \method and scored on their decoded frames.\footnote{AdaJEPA's update recipe, one gradient step on the last predictor block every five transitions, did not improve its predictions (\cref{app:baselines}).}

\subsection{Predicting the Mechanics of Unseen Games}
\label{sec:curves}

We ask how many exposures each model needs before it predicts the mechanics of an unseen game.
Reached@$k$ answers this directly as the share of mechanics that have reached the criterion by exposure $k$ (\cref{fig:cumulative}a).
LCA@$k$ complements it with the accuracy averaged over exposures up to $k$ and over all mechanics, including those that have not reached the criterion (\cref{fig:cumulative}b).

\method brings half of the mechanics to criterion within 36 exposures and 62\% within 100, more than twice the share of any baseline.
Its LCA@$k$ is the highest of all models after the first few exposures.
Frozen \method brings few mechanics to criterion, and its LCA@$k$ stops increasing after a few exposures, so this speed comes from learning at test time.
Updating the weights is not sufficient, however, since AdaJEPA and DreamerV3 are also updated after every transition and remain low in both panels.
Qwen3.8 is the strongest baseline.
It improves steadily as its context grows but remains far below \method.
Episodic retrieval brings no further mechanics to criterion after about 100 exposures.

\subsection{Memorization or Generalization}
\label{sec:generalization}

\begin{figure}[t]
  \centering
  \includegraphics[width=\linewidth]{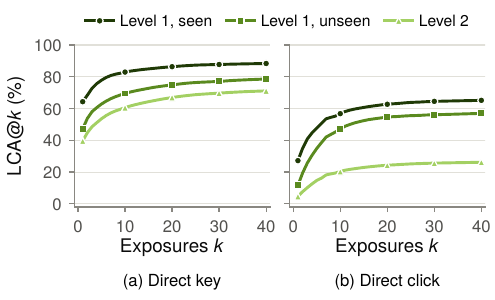}
  \caption{\textbf{Generalization to other recordings.}
  After $k$ exposures to a mechanic on level 1 of one recording, \method predicts that mechanic in the game's other recordings without further updates.
  Curves show LCA@$k$, averaged over exposures 1 to $k$, for seen and unseen transitions in level 1 and for transitions in level 2.}
  \label{fig:generalization}
\end{figure}

The accuracy gains in \cref{sec:curves} could reflect memorizing transitions rather than learning the mechanic.
We test this on \emph{direct mechanics}, in which a key press or click changes the object it acts on.
We adapt \method on level 1 of one recording and, after $k$ exposures to a mechanic, predict that mechanic in levels 1 and 2 of the game's other recordings without further updates.
A level-1 transition is seen if the adapted recording has already shown the same key acting at the same place or a click on the same object at the same cell, and unseen otherwise (\cref{app:novelty}).

\paragraph{Seen and unseen transitions.}
\Cref{fig:generalization} shows that \method also predicts unseen transitions.
In level 1, LCA@$k$ rises with $k$ on unseen transitions as well as on seen ones, and the gap between the two is similar for keys and clicks.
In level 2, LCA@$k$ is lower than on unseen level-1 transitions, slightly for keys and much more for clicks.

\begin{figure}[t]
  \centering
  \includegraphics[width=\linewidth]{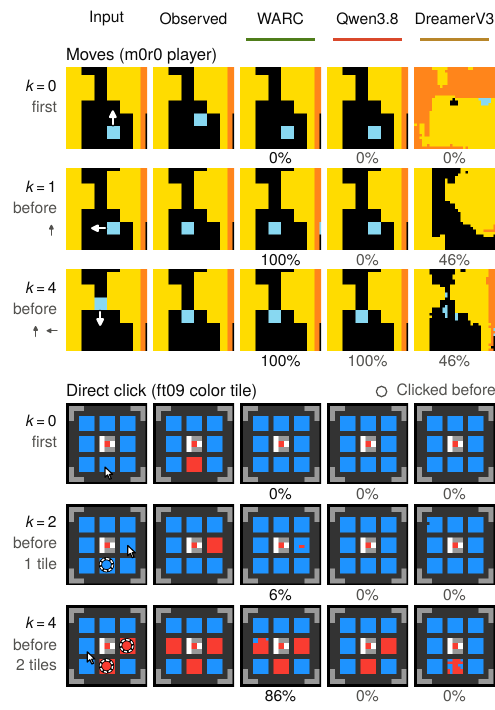}
  \caption{\textbf{Generalization to unseen transitions.}
  Each mechanic is shown at its first exposure ($k=0$) and at its first two exposures in a new direction or on a new tile, with what the recording has shown before at the left.
  Percentages are the share of changed cells predicted correctly.}
  \label{fig:examples}
\end{figure}

\paragraph{Kinds of unseen transitions.}
An unseen move is either a known move from a new position or a move in a new direction, and an unseen click is either on a clicked object at a new cell or on a new object.
Known and new refer to what the adapted recording has shown before.
\Cref{fig:examples} shows cases in which \method predicts a new direction or object after a few exposures, the first move to the left after one move up and a click on a third tile after clicks on two other tiles.
Over all moves and clicks, however, the new direction or new object is predicted less well than the new position or new cell (\cref{app:novelty}).
\method therefore generalizes beyond the transitions it has seen, also to new directions and objects, but less reliably than to known changes at new places.

\begin{table}[t]
  \centering
  \footnotesize
  \setlength{\tabcolsep}{6pt}
  \begin{tabular}{@{}lrrrr@{}}
    \toprule
     & LCA@10$\uparrow$ & LCA@100$\uparrow$ & n.r.$\downarrow$ & FE$\downarrow$ \\
    \midrule
    \textbf{\method} & \textbf{53.1} & \textbf{73.8} & \textbf{31.4} & \textbf{0.11} \\
    \midrule
    Frozen \method & 17.5 & 18.4 & 95.7 & 0.11 \\
    Recurrent \method & 21.5 & 23.8 & 85.5 & 0.24 \\
    From-scratch \method & 16.4 & 36.8 & 75.8 & 1.15 \\
    \bottomrule
  \end{tabular}
  \caption{Comparison of \method and its variants on \evalsuite. Recurrent \method uses a recurrent state instead of weight updates. LCA@10 and LCA@100 are the learning curve area at exposures 10 and 100. n.r. is the share of mechanics that do not reach the criterion. FE is the false-edit rate. All values are percentages.}
  \label{tab:etc}
\end{table}

\subsection{What Makes a World Model Learn Fast}
\label{sec:ablations}

\method adapts to a new game with two sources, prior knowledge from pretraining on other games and new experience from a gradient update after each transition.
We ask how much each source contributes and how the update should be made.
We ablate pretraining and test-time updates separately, replace updates with a recurrent state, and vary the update procedure.

\paragraph{Test-time updates.}
\Cref{tab:etc} shows that frozen \method, which is not updated at test time, reaches about a quarter of the LCA@100 of \method and is below it in every fold and test-time seed.
Recurrent \method, which reads the recording through a convolutional GRU instead of being updated, is almost as low.
When the transitions in which a mechanic fires are withheld from the replay buffer, with the number of updates unchanged, its LCA$_m$@$k$ stays close to that of frozen \method (\cref{app:ablations}).
Updates on the rest of the recording thus do little for that mechanic.

\paragraph{Pretraining.}
Pretraining sets how fast \method learns to predict a new mechanic.
From-scratch \method, which starts from random initialization, learns from the same exposures, reaches about half the LCA@100 of \method, and brings far fewer mechanics to criterion.

\begin{table}[t]
  \centering
  \footnotesize
  \setlength{\tabcolsep}{8pt}
  \begin{tabular}{@{}lrr@{}}
    \toprule
     & LCA@10 & LCA@100 \\
    \midrule
    \method & 53.1 & 73.8 \\
    \midrule
    Update every 5 transitions & 49.6 & 72.6 \\
    \multicolumn{3}{@{}l}{\emph{Updated parameters}} \\
    \quad LoRA, rank 64 & 55.3 & 72.2 \\
    \quad Last two blocks & 47.2 & 69.5 \\
    \quad Modulation layers & 43.5 & 63.9 \\
    \bottomrule
  \end{tabular}
  \caption{Ablations of the test-time update of \method on \evalsuite. LCA@10 and LCA@100 as in \cref{tab:etc}.}
  \label{tab:ablations}
\end{table}

\paragraph{Update frequency.}
With the total test-time compute budget held fixed, updating every five transitions reduces performance compared with updating after every transition (\cref{tab:ablations}).
The drop is larger for LCA@10 than for LCA@100, indicating that early learning is more sensitive to update frequency.

\paragraph{Updated parameters.}
The update need not change every parameter.
A low-rank update with LoRA~\cite{hu2022lora} in every block has a slightly lower LCA@100 than the full update and a higher LCA@10.
Updating only the modulation layers or only the last two transformer blocks lowers both.

\section{From Prediction to Play}
\label{sec:closedloop}

\evalsuite measures how accurately a world model predicts
the effects of individual mechanics.
However, improvements in prediction do not necessarily
translate into better decisions.
We therefore investigate whether test-time adaptation
of a world model improves exploration efficiency.
To this end, we integrate \method into a fixed graph explorer,
which then ranks candidate actions by their predicted outcomes.
We call the resulting agent \method-Agent.

\begin{figure*}[t]
  \centering    
  \makebox[\textwidth][c]{\hspace*{-0.66in}\includegraphics{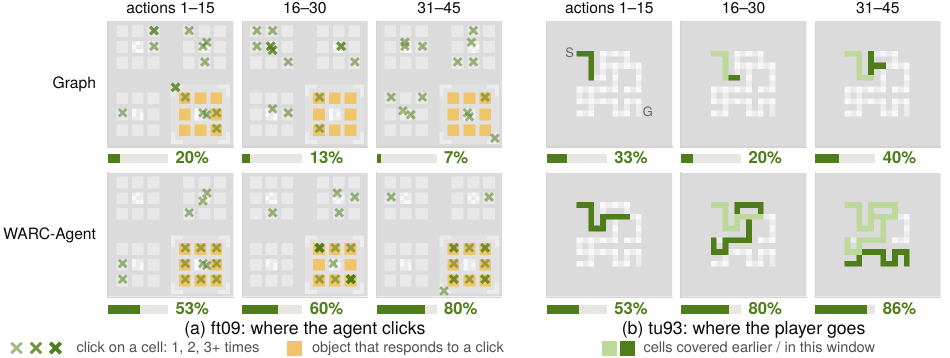}}
  \caption{\textbf{Where the explorer acts in ft09 and tu93.}
  In game ft09, a click changes the color of a tile; in game tu93, the keys move a player through a maze.
  The first 45 actions of level 1 in windows of 15, for Graph and \method-Agent.
  Bars show the share of actions that reached a game state not seen before in the level.}
  \label{fig:boards}
\end{figure*}

\begin{table}[t]
  \centering
  \footnotesize
  \setlength{\tabcolsep}{4pt}
  \begin{tabular}{@{}lccc@{}}
    \toprule
    Agent & L1@1000$\uparrow$ & L2@1000$\uparrow$ & RHAE@1000$\uparrow$ \\
    \midrule
    Random & 8.5\std{4.1} & 0.0\std{0.0} & 0.03\std{0.07} \\
    From-scratch CNN~\cite{smit2025goose} & 17.5\std{4.9} & 0.0\std{0.0} & 0.02\std{0.02} \\
    \midrule
    Graph & 30.0\std{4.7} & 7.0\std{4.2} & 0.17\std{0.10} \\
    Graph + frozen \method & 32.0\std{4.8} & 5.5\std{3.7} & 0.21\std{0.11} \\
    \method-Agent & \textbf{46.0}\std{4.6} & \textbf{10.5}\std{1.6} & \textbf{0.52}\std{0.25} \\
    \midrule
    Graph + oracle world model & 61.0\std{4.6} & 30.0\std{5.3} & 1.44\std{0.23} \\
    \bottomrule
  \end{tabular}
  \caption{Closed-loop play on the 20 games of \evalsuite with equal weight per game. L1@1000 and L2@1000 are the percentages of runs clearing levels 1 and 2 within 1,000 actions. RHAE@1000 is the official action-efficiency score. Graph is the explorer alone, and \method-Agent is Graph with \method. The from-scratch CNN is StochasticGoose~\cite{smit2025goose} reinitialized at each level. The oracle world model returns the true next frame and serves as an upper bound rather than an agent. Values are means over 10 seeds $\pm$ the standard deviation over seeds.}
  \label{tab:closedloop}
\end{table}

\subsection{Setup}
\label{sec:closedloop_setup}

\paragraph{Agent.}
The explorer is a simplified version of a published graph explorer~\cite{rudakov2025graph}, which we call Graph.
It hashes frames into states, records the actions tried in each state, and returns along known paths to states with untried actions.
A click can target any of the $64{\times}64$ cells, so the explorer narrows the click actions with heuristics (\cref{app:agent}).
Graph picks an untried action at random.
In \method-Agent, \method instead ranks the untried actions by the novelty of their predicted outcomes, as in model-based exploration~\cite{sekar2020plan2explore}.
\method predicts the next frame for each untried action and extends the most novel ones by a second step, and the agent takes the action whose predicted frame differs most from the frames seen in the level.

\paragraph{Protocol.}
Each agent plays each of the 20 games of \evalsuite with 10 random seeds.
The other five public games are left out, since they were used to choose the test-time recipe.
As in \cref{sec:experiments}, each game is played with the model of the fold that holds it out, so \method only plays games that it has not seen in pretraining.
All measures use the first 1{,}000 actions of each run.
We report the share of runs that clear levels 1 and 2, and the official score, Relative Human Action Efficiency (RHAE)~\cite{arcprize2026methodology}, on a 0--100 scale.

\subsection{Results}
\label{sec:closedloop_results}

\Cref{tab:closedloop} shows that \method-Agent triples the RHAE of Graph, from 0.17 to 0.52.
Frozen \method raises it only to 0.21, so most of the gain comes from learning at test time.
The same explorer with an oracle world model, which returns the true next frame, reaches nearly three times the RHAE of \method-Agent.
A world model that learns the mechanics faster would therefore make the explorer more efficient.

\Cref{fig:boards} visualizes how \method changes exploration behavior.
In ft09, Graph keeps clicking tiles that do nothing, while \method-Agent clicks the tiles that respond.
In tu93, Graph keeps to the corridors near the start, while \method-Agent takes the player through new parts of the maze to the goal.

\section{Discussion}
\label{sec:discussion}

\paragraph{Learning is still slow.}
Although WARC eventually reaches the criterion for two in
three mechanics, only one in five does so within ten exposures
(\cref{sec:curves}).
This suggests substantial room for improving the speed.
For perspective, people can learn to play unfamiliar video
games within minutes or a few hundred actions
~\cite{tsividis2017human,tsividis2021theory}.
One possible route to faster learning is in-context learning
without gradient updates.
Our LLM baseline performed moderately despite its
in-context learning capabilities, possibly because its
pretraining provides limited prior knowledge of game dynamics.
Similarly, a recurrent state did not replace gradient updates
in WARC (\cref{sec:ablations}).
Whether extensive pretraining on diverse games can enable
effective in-context learning of mechanics remains open.

\paragraph{Generalization is uneven across changes.}
A model that has learned a mechanic as a reusable rule
should predict its effects in unseen situations.
\method does so only in part.
It generalizes to unseen transitions, but moves in new
directions and clicks on new objects are predicted less
reliably than known changes at new places
(\cref{sec:generalization}).
Rotation augmentation improves generalization to new
directions, but does not close the gap.
These results suggest that learning reusable mechanics
requires more than spatial generalization, motivating
architectures or training methods that transfer learned
effects across different actions and objects.

\paragraph{Play needs more than a world model.}
ARC-AGI-3 asks an agent to explore, model the environment, infer goals, and plan~\cite{arcprize2026techreport}, while \evalsuite measures only world modeling.
The condition for completing a level is itself a mechanic, but it fires only once per level and \evalsuite does not score it.
Goal inference poses a different challenge from learning ordinary mechanics.
While some mechanics reach the criterion within a few exposures, the main difficulty in goal inference is identifying the condition that triggers completion, rather than predicting its effect.
An agent must infer this condition from a single successful example in level 1, distinguish it from incidental features of the state, and transfer it to the new layout of level 2.
Previous work on ARC-AGI-3 also finds such goal inference difficult~\cite{skoutnev2026twin,lehmann2026tycho,ahmad2026competence}.
Even with an oracle world model, our explorer clears no level in four of the 20 games (\cref{app:agent}).
Higher scores therefore require more than faster learning of mechanics, particularly the ability to infer goal conditions from sparse experience and plan toward satisfying them.

\paragraph{Limitations.}
\evalsuite covers levels 1 and 2 of the 20 public games whose changes can be attributed to single objects, and its annotator needs the game program or an environment that logs which mechanics fire.
It scores a mechanic only on the transitions of the human recordings, so reaching the criterion is weaker than learning the rule in general.
In closed-loop play, the action selector is hand-designed and shared across conditions, so our claims concern the world model rather than the agent.
All results come from one benchmark of deterministic, discrete games on a $64{\times}64$ grid, and updating after each transition is untested on continuous or stochastic dynamics.

\section{Conclusion}
\label{sec:conclusion}

We asked how many exposures a world model needs to predict the mechanics of a new game.
On unseen games, \method brings 62\% of all mechanics to criterion within a hundred exposures.
The speed comes from pretraining on other games and from a gradient update after each transition.
In closed-loop play, adding \method to a fixed explorer triples its efficiency score.
\evalsuite enables systematic comparison of next-frame models by measuring how quickly they learn to predict individual mechanics across 20 ARC-AGI-3 games. Its evaluation protocol can also be extended to other environments where mechanic activations and their effects can be identified.
We believe that world models for open-ended environments will be judged less by what they know after pretraining than by how fast they learn what they do not yet know.

\section*{Acknowledgements}
We thank Masahiro Suzuki and Shohei Taniguchi for their helpful comments on the manuscript.

\section*{LLM Usage Statement}
Claude Code and Claude Fable (Anthropic), along with Codex and GPT-6 (OpenAI), were used to draft and edit text and check grammar.
Claude Code was also used to write code for the experiments and check references.
OpenAI Codex was used to draft layouts for the concept figures, which the authors redrew using their own scripts.
The authors remain responsible for the scientific claims, results, citations, and final manuscript.

{
    \small
    \bibliographystyle{ieeenat_fullname}
    \bibliography{main}
}

\clearpage
\onecolumn
\raggedbottom   %
\begin{center}
  \Large\textbf{\thetitle}\\[0.5em]
  Appendix
\end{center}
\vspace{0.5em}
\suppressfloats[t]   %
\appendix
\crefalias{section}{appendix}
\renewcommand{\floatpagefraction}{0.4}
\makeatletter\setlength{\@fptop}{0pt}\setlength{\@fpsep}{8pt}\makeatother   %

\section{\evalsuite Details}
\label{app:mecheval}

\paragraph{Games and recordings.}
\evalsuite covers 20 of the 25 public games, the same 20 as the concurrent study~\cite{concurrent2026perception}.
In the other five, the changed cells cannot be attributed to single objects of the game program.
\Cref{tab:games} gives the data per game.

\begin{figure}[p]
  \centering
  \includegraphics[width=\linewidth]{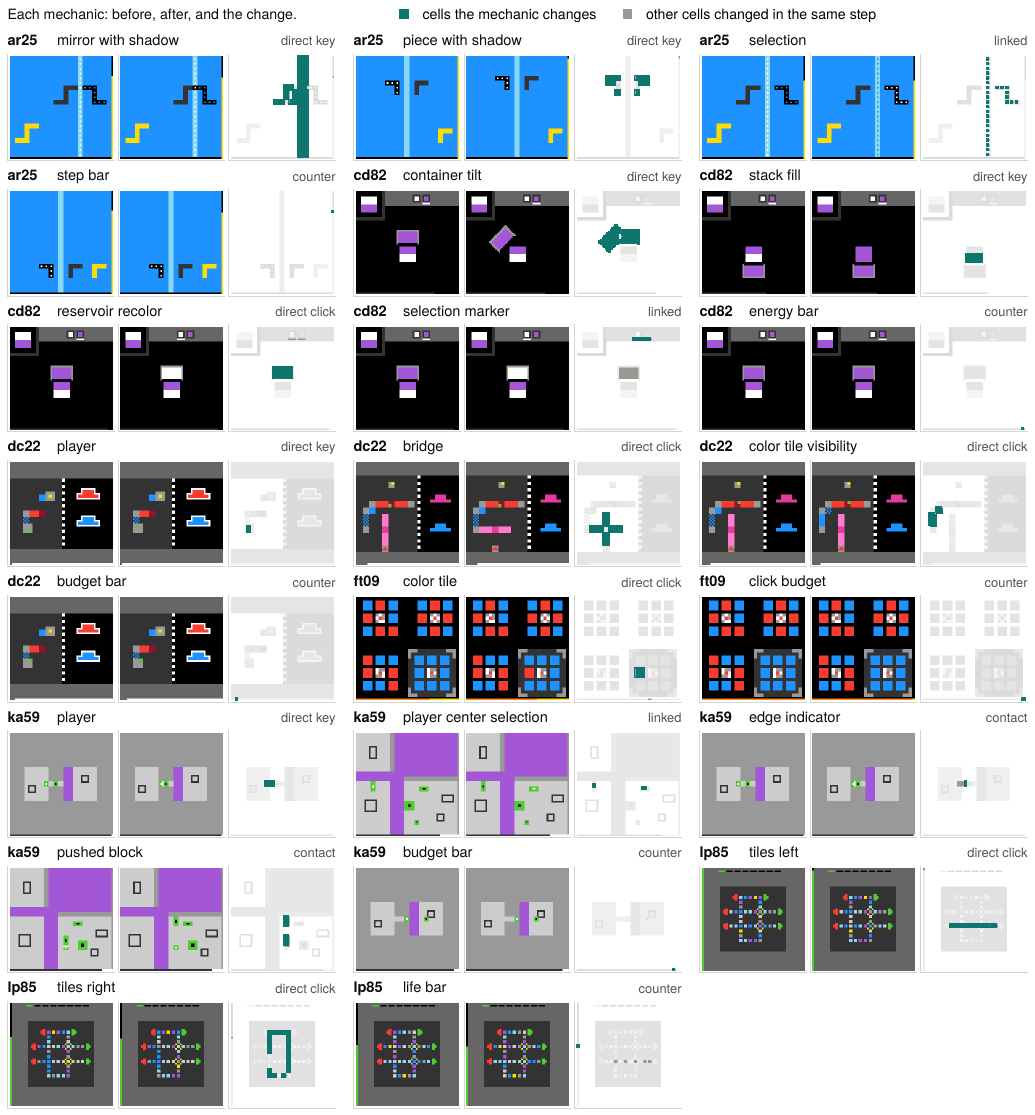}
  \caption{\textbf{The mechanics of \evalsuite, game by game: ar25 to lp85} (ls20 to wa30 in \cref{fig:mech_examples2,fig:mech_examples3}).
  Each mechanic is shown at one representative exposure, with its kind above the panel (\cref{tab:mech_kinds}).}
  \label{fig:mech_examples}
\end{figure}

\begin{figure}[p]
  \centering
  \includegraphics[width=\linewidth]{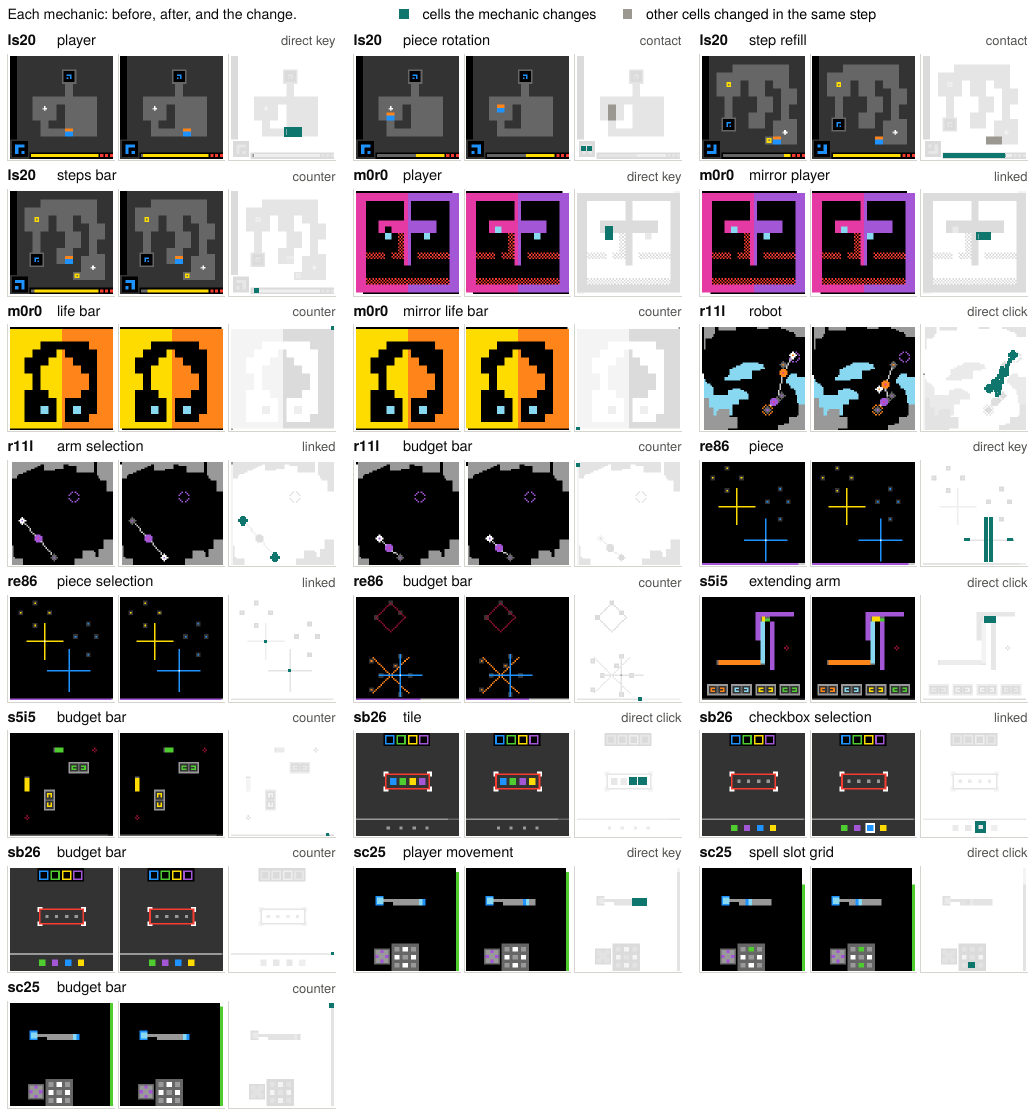}
  \caption{\textbf{The mechanics of \evalsuite, game by game: ls20 to sc25}, shown as in \cref{fig:mech_examples}.}
  \label{fig:mech_examples2}
\end{figure}

\begin{figure}[p]
  \centering
  \includegraphics[width=\linewidth]{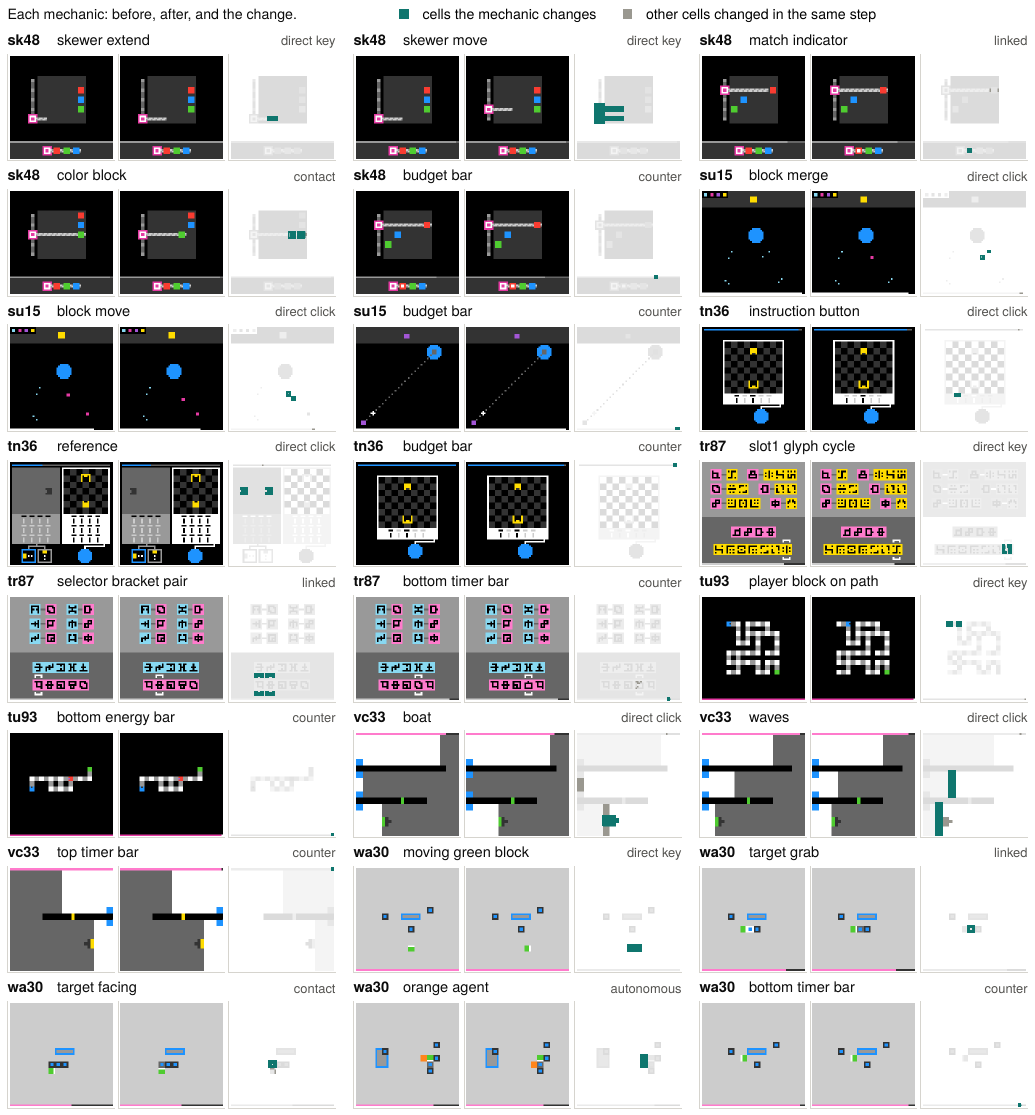}
  \caption{\textbf{The mechanics of \evalsuite, game by game: sk48 to wa30}, shown as in \cref{fig:mech_examples}.}
  \label{fig:mech_examples3}
\end{figure}

\paragraph{Mechanics.}
A mechanic is a branch of the game program that updates the state, defined by its condition, the object it changes, and its effect.
Parts of the game state that always change together form one object, and a change to another object is a separate mechanic.
We started from the list of mechanics of the concurrent study~\cite{concurrent2026perception} and checked each entry against the programs and the recordings.
\chg{\suppcopynote}
We removed entries whose objects do not appear in levels 1 and 2 or whose change is visible only during an animation, merged or split entries to follow the definition above, and added changes that the list had missed.
This gives 81 mechanics.
Three of them never fire in the recordings, two never change a visible cell, and seven fire fewer than three times in the median recording that contains them (\cref{sec:mecheval}).
Examples of the seven are picking up a key and the flash at the goal.
The remaining 69 mechanics are scored.
\Cref{fig:mech_examples,fig:mech_examples2,fig:mech_examples3} show every mechanic at one exposure.

\paragraph{Kinds.}
\Cref{tab:mech_kinds} defines the six kinds and their three groups.
Two machine learning researchers sorted the mechanics into the three groups.
They did not see our kinds.
Our groups agree with each rater on 86\% of the mechanics (Cohen's $\kappa$ 0.77), and the two raters agree with each other on 87\% ($\kappa$ 0.78).

\paragraph{Exposures and scored cells.}
Mechanic $m$ fires at step $t$ if its branch runs and the state of its object changes.
When an action plays an animation, we use the state after it has ended.
The scored cells $C_m(t)$ of $m$ are the changed cells inside the bounding boxes of its objects, taken before and after the step.
When the program's log does not name every changed object, or the boxes of two mechanics overlap, fixed rules assign the cells.
For example, the mechanic of a selection marker is scored only on the marker's cells, and the cells of the selected object are scored under that object's mechanic.
After these rules, 0.03\% of the changed cells are assigned to no mechanic.
Steps where a mechanic's condition fails, such as a move blocked by a wall, are not exposures to that mechanic.
They are recorded and counted separately (\cref{app:novelty}).
The two raters independently checked the scored cells on four sampled exposures of every mechanic.
They agreed on 94\% of the exposures, and 93\% of all judgments found the cells correct.
After the two raters had finished, we revised the mechanics that either of them flagged.

\begin{table}[tp]
  \centering
  \footnotesize
  \begin{tabular}{@{}lclrrrrr@{}}
    \toprule
    Game & Fold & Actions & Recordings & Transitions & Mechanics & Exposures & Exposures per mechanic \\
    \midrule
    ar25 & 3 & keys, click & 10 & 1{,}002 & 4 & 1{,}570 & 32 \\
    cd82 & 2 & keys, click & 11 & 654 & 5 & 824 & 7 \\
    dc22 & 2 & keys, click & 11 & 1{,}632 & 4 & 2{,}064 & 22.5 \\
    ft09 & 3 & keys, click & 10 & 547 & 2 & 636 & 28 \\
    ka59 & 3 & keys, click & 10 & 1{,}163 & 5 & 1{,}611 & 16 \\
    lp85 & 1 & click & 54 & 4{,}266 & 3 & 6{,}834 & 24 \\
    ls20 & 3 & keys & 13 & 2{,}411 & 4 & 4{,}745 & 35 \\
    m0r0 & 2 & keys, click & 11 & 1{,}748 & 4 & 4{,}169 & 88 \\
    r11l & 1 & click & 10 & 533 & 3 & 985 & 27 \\
    re86 & 1 & keys & 11 & 1{,}008 & 3 & 1{,}653 & 41 \\
    s5i5 & 3 & click & 11 & 1{,}942 & 2 & 2{,}460 & 62 \\
    sb26 & 1 & keys, click & 12 & 824 & 3 & 973 & 22.5 \\
    sc25 & 1 & keys, click & 15 & 694 & 3 & 865 & 13 \\
    sk48 & 2 & keys, click & 14 & 3{,}730 & 5 & 6{,}280 & 66 \\
    su15 & 1 & click & 13 & 844 & 3 & 1{,}262 & 27 \\
    tn36 & 2 & click & 14 & 1{,}462 & 3 & 2{,}351 & 34 \\
    tr87 & 2 & keys & 12 & 1{,}696 & 3 & 2{,}539 & 64.5 \\
    tu93 & 2 & keys & 13 & 492 & 2 & 938 & 32.5 \\
    vc33 & 3 & click & 10 & 325 & 3 & 659 & 21 \\
    wa30 & 1 & keys & 14 & 2{,}920 & 5 & 6{,}178 & 62 \\
    \midrule
    All & & & 279 & 29{,}893 & 69 & 49{,}596 & 30 \\
    \bottomrule
  \end{tabular}
  \caption{\evalsuite data per game, up to the first completion of level 2. Fold is the fold whose model is not pretrained on the game. Exposures per mechanic is the median number of exposures over all pairs of a mechanic and a recording in which it fires.}
  \label{tab:games}
\end{table}

\begin{table}[tp]
  \centering
  \footnotesize
  \begin{tabular}{@{}lllcl@{}}
    \toprule
    Group & Kind & Definition & Mechanics & Example \\
    \midrule
    Direct & Direct key & A key moves, tilts, extends, or changes an object & 15 & ls20 player moves \\
     & Direct click & A click changes the clicked object or the object that a button drives & 16 & lp85 tile row shifts \\
    \addlinespace
    Other & Linked & Copies or marks another change & 10 & re86 piece selection \\
     & Contact & An object reaches a place or touches another object & 6 & ka59 pushed block slides \\
     & Autonomous & An object moves by its own rule & 1 & wa30 orange agent \\
    \addlinespace
    Counter & Counter & A step or budget bar changes & 21 & ls20 steps bar \\
    \bottomrule
  \end{tabular}
  \caption{The six kinds of the 69 mechanics, in three groups.}
  \label{tab:mech_kinds}
\end{table}


\section{\method Details}
\label{app:model}

This section gives the details that \cref{sec:method} omits.
\Cref{tab:varc} compares \method with VARC, and \cref{tab:hparams} lists the training settings.

\begin{table}[t]
  \centering
  \footnotesize
  \begin{tabular}{@{}p{0.14\linewidth}p{0.38\linewidth}p{0.42\linewidth}@{}}
    \toprule
    & VARC & \method \\
    \midrule
    \multicolumn{3}{@{}l}{\emph{Model}} \\
    Task & output grid of a static puzzle & next frame of an interactive game \\
    Input & one grid on a $64{\times}64$ canvas & the last three $64{\times}64$ frames and actions \\
    Transformer & width 512, 10 blocks, 8 heads, $2{\times}2$ patches & the same, with patches that span the three frames \\
    Output & a color for every cell & the same \\
    Conditioning & one learned token per puzzle & the actions, through adaptive layer normalization and a learned vector at the clicked patch \\
    Parameters & 18M & 35.3M \\
    \midrule
    \multicolumn{3}{@{}l}{\emph{Pretraining}} \\
    Data & the 400 ARC training puzzles and RE-ARC, about 400k pairs & human recordings of 16 or 17 games, 90k to 123k transitions \\
    Augmentation & scale and translation on the canvas & rotations, flips, and color permutations, 51 copies \\
    \midrule
    \multicolumn{3}{@{}l}{\emph{Test-time training}} \\
    Data & the 2 to 4 demonstration pairs of the puzzle & the transitions observed so far \\
    Schedule & 100 epochs before the prediction & 20 steps after every transition, which is predicted before training on it \\
    Augmentation & flips, rotations, and color permutations, 51 copies, each with its own token & rotations only, 31 copies \\
    Prediction & majority vote over 510 views & one forward pass \\
    \bottomrule
  \end{tabular}
  \caption{What \method takes from VARC and what it changes (\cref{sec:method}). The VARC column follows \citet{hu2025varc} for their default model.}
  \label{tab:varc}
\end{table}
\paragraph{Architecture.}
Each input cell is one of 16 colors.
History frames that fall before the start of an episode or level are filled with a padding symbol.
A 3D convolution with kernel and stride $3{\times}2{\times}2$ turns the three embedded frames into $32{\times}32$ tokens with rotary position embeddings, followed by the transformer of VARC (\cref{tab:varc}).
The current action is embedded by its type and, for a click, by its $x$ and $y$ coordinates, and conditions every block through adaptive layer normalization initialized at zero~\cite{peebles2023dit}.
A click also adds a learned vector to the token of the clicked patch.
There are four such vectors, one for each position of the clicked cell within its $2{\times}2$ patch.
The two earlier actions are embedded in the same way, concatenated, and added to the condition of the adaptive layer normalization through a linear layer initialized at zero.
The output head predicts the $2{\times}2$ cells of each token.
\method has 35.3M parameters, of which the adaptive layer normalization holds 15.8M and the patch embedding 3.1M.

\providecommand{\tttcompute}{one GPU}
\begin{table}[t]
  \centering
  \footnotesize
  \setlength{\tabcolsep}{4pt}
  \begin{tabular}{@{}lll@{}}
    \toprule
    & Pretraining & Test-time training \\
    \midrule
    Optimizer & Muon + AdamW & Muon + AdamW \\
    Learning rate & $3{\times}10^{-4}$ & $3{\times}10^{-4}$ \\
    Weight decay & 0 & 0.01 \\
    Schedule & warmup, cosine & constant \\
    Batch & 516 & 8 \\
    Length & 30 epochs & 20 steps per transition \\
    Augmented copies & 51 & 31 (rotations) \\
    Precision & bfloat16 & bfloat16 \\
    Gradient clipping & 1.0 & 1.0 \\
    Compute & 13--20\,h per fold on six GPUs & \tttcompute \\
    \bottomrule
  \end{tabular}
  \caption{Training settings.}
    \label{tab:hparams}
\end{table}
\paragraph{Pretraining.}
The 25 public games are split at random into three folds, with the six games without clicks and the five games outside \evalsuite spread evenly over the folds.
\Cref{tab:games} gives the fold of each game of \evalsuite, and the other five fall in fold 1 (bp35, lf52), fold 2 (cn04), and fold 3 (g50t, sp80).
We train each fold model on the human recordings of its 16 or 17 training games, 90k to 123k transitions after removing resets.
A reset is an action that restarts a level or the game.
For each training game, one recording that reaches the last level is held out to select the checkpoint by exact-frame accuracy, the fraction of next frames predicted with every cell correct.
Each recording is augmented into 51 copies.
They are the original and 50 transformed copies, each one of three rotations or two flips combined with the original colors or one of nine color permutations.
The actions are transformed with the frames.

\paragraph{Test-time training.}
A transition is valid unless it crosses a reset, a level change, or the end of an episode, and the history restarts at these boundaries.
After every valid transition, the model takes 20 gradient steps.
Each minibatch of eight is drawn from a pool of the ten most recent valid transitions and twenty older ones.
The older ones are drawn at random once per update.
Each of these transitions enters as 31 copies, the original and each of its three rotations ten times, so each rotation is drawn ten times as often as the original.
The click positions are rotated with the frames and the four movement keys are relabeled, which assumes that these keys move in the four directions.
The optimizer state persists across updates, and every parameter of the model is updated.

\paragraph{Optimization.}
In both stages, Muon~\cite{jordan2024muon} updates the two-dimensional weights of the transformer blocks except the adaptive layer normalization, with momentum 0.95, and AdamW~\cite{loshchilov2019adamw} updates the remaining parameters.
Pretraining warms up the learning rate linearly over six epochs and then decays it with a cosine schedule~\cite{loshchilov2017sgdr}.
All runs use RTX PRO 6000 GPUs.

\paragraph{Choosing the recipe.}
We chose the test-time recipe on the five public games outside \evalsuite (61 recordings), each with the fold model that holds it out, by the following rule.
The score is the accuracy on the changed cells of every valid transition that changes the frame, predicted before the update on that transition.
Candidates whose false-edit rate exceeds twice the lowest are discarded.
We started from AdamW at $10^{-4}$ with five recent and ten older transitions and rotations with color permutations, and changed one setting at a time.
This gave Muon at $3{\times}10^{-4}$, ten recent and twenty older transitions, and rotations only.
The final recipe scores 82.8\% over three seeds, against 82.0\% for the starting point.

\begin{figure}[t]
  \centering
  \includegraphics[width=0.6\linewidth]{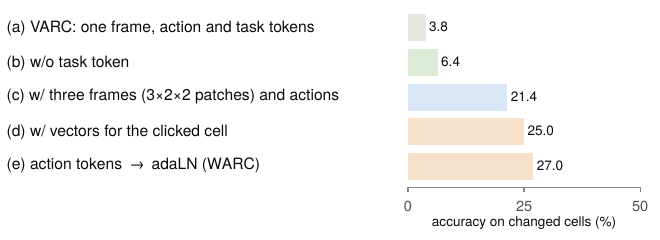}
  \caption{\textbf{From VARC to \method.} \chg{Accuracy on the changed cells of the seven unseen games of fold 2 in \evalsuite after pretraining, before any update. Each row adds one change to the row above. One pretraining run per row.}}
  \label{fig:varc_to_warc}
\end{figure}

\paragraph{From VARC to \method.}
\chg{\Cref{fig:varc_to_warc} starts from VARC applied to next-frame prediction and adds the changes of \cref{sec:model} one at a time on fold 2.
The first row sees one frame and gets the action and a task token as extra tokens before the patch tokens.
Each training game has its own task token, and a game outside them uses their mean.
From the third row on, each of the three action tokens also gets a learned embedding of its position in the history.
All rows use the same pretraining recipe.
Accuracy on the unseen games rises with every change, most with the last three frames.}

\FloatBarrier
\section{Baselines and Recurrent \texorpdfstring{\method}{WARC}}
\label{app:baselines}

AdaJEPA, DreamerV3 and recurrent \method are pretrained on the same folds as \method. AdaJEPA and DreamerV3 are updated at test time and take 20 gradient steps on minibatches of eight after every valid transition, and recurrent \method is not updated. Qwen3.8 is used as released and is not updated.
\Cref{tab:baselines} gives the results of the baselines.

\begin{table}[t]
  \centering
  \footnotesize
  \setlength{\tabcolsep}{3.5pt}
  \begin{tabular}{@{}lccccc@{}}
    \toprule
    Method & $k$=0$\uparrow$ & LCA@10$\uparrow$ & LCA@100$\uparrow$ & n.r.$\downarrow$ & FE$\downarrow$ \\
      & \% & \% & \% & \% & \% \\
    \midrule
    AdaJEPA & 2.0 & 12.5 & 22.2 & 91.3 & 1.41 \\
    DreamerV3 & \textbf{14.2} & 22.2 & 27.1 & 98.6 & 3.07 \\
    Episodic retrieval & 0.1 & 24.7 & 34.4 & 85.5 & \textbf{0.08} \\
    Qwen3.8 & 0.8 & 31.1 & 46.7 & 73.9 & 0.31 \\
    \midrule
    \method & 3.7 & \textbf{53.1} & \textbf{73.8} & \textbf{31.4} & 0.11 \\
    \bottomrule
  \end{tabular}
  \caption{The baselines of \cref{sec:setup} on \evalsuite, with the columns of \cref{tab:etc} and the share of changed cells predicted correctly before any exposure ($k$=0). \method, AdaJEPA, and DreamerV3 average three test-time seeds, and Qwen3.8 is run once. Bold: best per column.}
  \label{tab:baselines}
\end{table}

\paragraph{AdaJEPA.}
We use the released code of AdaJEPA~\cite{wang2026adajepa}.
A convolutional encoder maps each frame to latent tokens, and a transformer predicts the tokens of the next frame from those of three frames and the actions.
The loss is the squared error to the encoder's tokens of the next frame, with a stop-gradient on the target.
A decoder is trained on the latents and passes no gradient to the encoder or the predictor.
We match the inputs and outputs to \method.
One token covers $2{\times}2$ cells, the clicked cell is marked in the input, and the decoder outputs one of the 16 colors per cell.
A model trained with the latent loss alone predicted few of the changed cells.
We therefore add an inverse-dynamics loss~\cite{sobal2025pldm}, which predicts the action from the latents of consecutive frames.
The optimizers and learning rates are those of the released code.
We keep the checkpoint with the best accuracy on changed cells of held-out recordings.
The model has 36.3M parameters.
At test time, every parameter is updated with the pretraining loss and the recipe of \method, and the decoder is trained on the observed frame.
The learning rate, $3{\times}10^{-4}$, was chosen on the five games outside \evalsuite, and the encoder keeps its pretraining learning rate.
The adaptation of the original method, one gradient step on the last predictor block every five transitions, did not improve the predictions.
The low accuracy comes from the predicted latents rather than the decoder.
Decoding the encoder's latents of the true next frames with the pretrained weights gives an LCA@100 of 80\% on \evalsuite, against 22\% for the predicted latents.
Where a predicted token is closer to the token of the next frame than to that of the current frame, the decoder renders all of its changed cells correctly in 78\% of cases, against 10\% otherwise.
A better-trained latent does not change this.
On fold 3, we continued the pretraining with the released recipe, halving the learning rate whenever the loss diverged, and took the checkpoint with the lowest latent error on held-out recordings of the training games, 14\% below that of the selected checkpoint.
On the six games of fold 3 in \evalsuite, it reaches an LCA@100 of 25\% after test-time training, the same as the selected checkpoint, and its predicted latents are no closer to those of the next frames.

\paragraph{Episodic retrieval.}
Episodic retrieval stores every valid transition of the recording so far and has no parameters.
For each cell, it predicts the most frequent next color among earlier cells with the same $3{\times}3$ neighborhood under the same action type, breaking ties by recency.
For a click, the earlier cell must also have the same position relative to the clicked cell, where all positions more than two cells away count as one.
A cell without a match is predicted unchanged.
A $5{\times}5$ neighborhood raises the accuracy by less than two points, and variants that copy from the single most similar earlier transition score lower than both.
The mechanics that episodic retrieval brings to criterion are mostly counters.

\paragraph{DreamerV3.}
We use a PyTorch port of the DreamerV3 world model~\cite{hafner2023dreamerv3} that matches the official JAX implementation.
We take its 50M configuration with a CNN depth of 64 (45.9M parameters).
A categorical decoder over the 16 colors lets it be scored cell by cell like \method, and the action is factored into its type and click coordinates.
It is pretrained with only the world-model loss for 50 epochs on batches of 16 sequences of 64 steps, at a constant learning rate of $3{\times}10^{-4}$ after 1{,}000 warmup steps.
We keep the checkpoint with the best accuracy over all cells of the next frame on held-out recordings.
The depth and the learning rate were chosen by this accuracy on fold 1.
At test time, it predicts from a state recomputed over the last 64 frames.
Every part of the model is updated.
For the update, each sampled transition is the last step of a sequence of up to nine frames.
The update of DreamerV3 was tuned on the five games outside \evalsuite by the rule used for \method.
The tuning gave AdamW at a constant $3{\times}10^{-4}$, five recent and ten older transitions, and rotations only.

\paragraph{Recurrent \method.}
Recurrent \method keeps the blocks, the click vectors, and the head of \method and replaces the three-frame history by a convolutional GRU over the $32{\times}32$ patch grid.
Each transition is encoded into a 64-channel map from its two frames, its action, and a flag for episode and level boundaries.
The GRU begins each recording from a zero state and reads these maps in order.
The last map encodes the current frame alone.
The state, mapped to the token width, forms the input tokens, so the model sees the frames only through the state.
The action also enters as a prefix token and through adaptive layer normalization.
The model has 32.2M parameters.
We pretrain one model per fold with the training settings of \method (\cref{tab:hparams}), reading each recording in chunks of 64 transitions with the state carried across chunks.

\paragraph{LLM (Qwen3.8).}
We query Qwen3.8-Flash (\texttt{qwen/qwen3.8-flash}), the hosted version of the open-weight Qwen3.8-Flash-Next~\cite{qwen2026flashnext}, through OpenRouter, served by Alibaba Cloud (\texttt{alibaba/us-east-1}), at temperature 0 and with reasoning off.
The system prompt and the user message are given below.
When a prompt would not fit the context of about one million tokens, the oldest transitions are dropped until it fills 90\% of it.
The whole recording so far fits in 85\% of the predictions, and the others keep the last 199 to 222 transitions.
Rows that the answer leaves out are copied from the current frame.
An answer with fewer than 32 readable rows is requested again up to five times, and one that is still unreadable counts as predicting no change (0.3\% of the predictions).
Each frame is written in full, as \method receives and predicts it.
On 200 transitions drawn at random from the five games outside \evalsuite, the model predicted 53.5\% of the changed cells correctly with reasoning on (a budget of 16k tokens) and 59.0\% with it off, and each call took four times as long with reasoning on.

\makeatletter
\newcommand{\llmneedspace}[1]{\par\penalty-100\begingroup\setlength{\dimen@}{#1}\dimen@ii\pagegoal\advance\dimen@ii-\pagetotal\ifdim\dimen@>\dimen@ii\ifdim\dimen@ii>\z@\vfil\fi\penalty-\@M\fi\endgroup}%
\makeatother
\llmneedspace{10\baselineskip}
\paragraph{System prompt.} It is the same for every call.\par\nopagebreak
\begin{tcolorbox}[breakable, title=System Prompt]
\small
You predict the next frame of an unknown turn-based game from a recording of a person playing it.

\vspace{0.4em}
A frame is a 64x64 grid of cells. Each cell holds one of 16 colors, written as a hexadecimal digit 0-f. A frame is written as 64 lines, top row first; each line starts with its row number 00-63 and a bar, followed by the 64 digits of that row, e.g.\ \texttt{"07|5555333..."}.

\vspace{0.4em}
An action is written by its identifier (ACTION1 to ACTION7). ACTION6 is a click at the cell (x, y), where x is the column and y the row, both counted from 0 at the top-left. RESET restarts the level.

\vspace{0.4em}
You are not told what the actions do or what the rules of the game are. Infer them from the transitions shown so far.

\vspace{0.4em}
Answer with the frame that follows the last action and nothing else: the 64 numbered rows 00-63, each with its 64 digits, in the same format, between \texttt{<frame>} and \texttt{</frame>}.
\end{tcolorbox}

\llmneedspace{12\baselineskip}
\paragraph{User message.} It holds the recording so far and grows by one transition at every step. In the template below, \texttt{<action\_i>} is an action written as in the system prompt, such as \texttt{ACTION6 x=12 y=40}, and \texttt{<frame\_i>} is a frame of 64 numbered rows. After the oldest transitions are dropped, the message starts with ``Frame before the transitions shown:'' instead.\par\nopagebreak
\begin{tcolorbox}[breakable, title=User Message (template)]
\small
Frame at the start of the recording:\\ \texttt{<frame\_0>}\\
Action: \texttt{<action\_1>}\\ Frame after it:\\ \texttt{<frame\_1>}\\
\hspace*{1.2em}$\vdots$\\
Action: \texttt{<action\_t>}\\ Frame after it:\\ \texttt{<frame\_t>}\\
Next action: \texttt{<action\_t+1>}\\
Write the frame that follows it.
\end{tcolorbox}

\llmneedspace{12\baselineskip}
\paragraph{Frame.} The frame below is the first frame of an ar25 recording, with rows 01 to 13 and 21 to 62 left out.\par\nopagebreak
\begin{tcolorbox}[breakable, title=Frame (excerpt)]
\scriptsize
\texttt{00|999999999999999999999999999999aaa999999999999999999999999999999b}\\
\hspace*{1.2em}$\vdots$\\
\texttt{14|999999999999999999999999999999aaa999999999999999999999999999999b}\\
\texttt{15|999999999999999999555555555999aaa999444444444999999999999999999b}\\
\texttt{16|999999999999999999505505505999aaa999444444444999999999999999999b}\\
\texttt{17|999999999999999999555555555999aaa999444444444999999999999999999b}\\
\texttt{18|999999999999999999999999555999aaa999444999999999999999999999999b}\\
\texttt{19|999999999999999999999999505999aaa999444999999999999999999999999b}\\
\texttt{20|999999999999999999999999555999aaa999444999999999999999999999999b}\\
\hspace*{1.2em}$\vdots$\\
\texttt{63|555555555555555555555555555555555555555555555555555555555555555b}
\end{tcolorbox}

\FloatBarrier
\section{Learning Curves}
\label{app:curves}
\suppressfloats[t]   %

\paragraph{All methods.}
The three test-time seeds of \method differ by at most 0.6 points in LCA@10 and 0.3 points in LCA@100.
Those of every other method and ablation that averages three test-time seeds differ by less than one point in both, so \cref{tab:etc,tab:ablations,tab:baselines} give means only.
In every fold and seed, the updates raise LCA@100 by 50 to 59 points over frozen \method.

\begin{figure}[t]
  \centering
  \includegraphics[width=\linewidth]{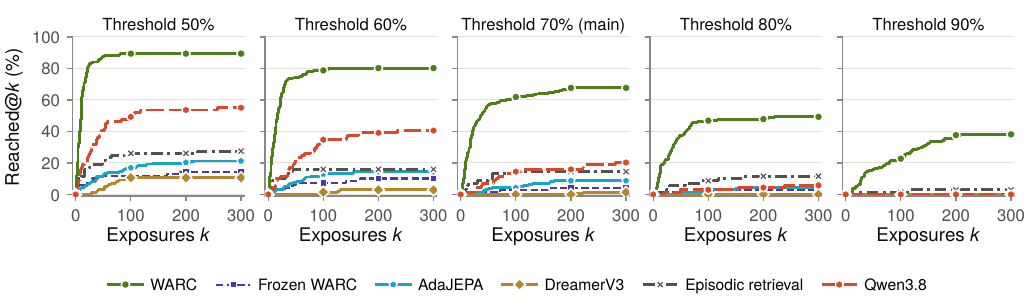}
  \caption{\textbf{Reached@$k$ under other thresholds.} Reached@$k$ as in \cref{fig:cumulative}a, with the threshold of the criterion at 50\%, 60\%, 70\% (the main criterion), 80\%, and 90\%. \method, AdaJEPA, and DreamerV3 average three test-time seeds, and Qwen3.8 is run once.}
  \label{fig:reached_tau}
\end{figure}

\paragraph{Sensitivity of the criterion.}
A model that is wrong on the first three of ten exposures and right afterwards reaches $\tau=70\%$ at the tenth exposure, and frozen \method, which does not update at test time, reaches it for 3 of the 69 mechanics.
\Cref{fig:reached_tau} shows Reached@$k$ under the thresholds of 50\%, 60\%, 70\%, 80\%, and 90\%.
Under each threshold, no other method brings more mechanics to criterion than \method at any exposure.
When games and then recordings are resampled 2{,}000 times, the share of \method under the main criterion has a 95\% bootstrap interval of 59\% to 77\%, and the gap to every other method has an interval that excludes zero.

\paragraph{\chg{Per recording.}}
\chg{The LCA of \cref{sec:mecheval} is computed on learning curves averaged over recordings.
\Cref{tab:per_recording} also computes it on the curve of each recording and then averages over the recordings, with equal weight per mechanic.
\method stays far ahead of every other method.}

\begin{table}[t]
  \centering
  \footnotesize
  \setlength{\tabcolsep}{5pt}
  \chg{\begin{tabular}{@{}lrrrr@{}}
    \toprule
     & \multicolumn{2}{c}{Averaged curve} & \multicolumn{2}{c}{Per recording} \\
    \cmidrule(lr){2-3}\cmidrule(l){4-5}
    Method & LCA@10 & LCA@100 & LCA@10 & LCA@100 \\
    \midrule
    \method & 53.1 & 73.8 & 49.7 & 63.2 \\
    Frozen \method & 17.5 & 18.4 & 17.6 & 18.4 \\
    Recurrent \method & 21.5 & 23.8 & 21.5 & 23.8 \\
    From-scratch \method & 16.4 & 36.8 & 15.2 & 27.2 \\
    AdaJEPA & 12.5 & 22.2 & 11.7 & 17.4 \\
    DreamerV3 & 22.2 & 27.1 & 21.3 & 24.2 \\
    Episodic retrieval & 24.7 & 34.4 & 23.1 & 30.3 \\
    Qwen3.8 & 31.1 & 46.7 & 29.2 & 40.3 \\
    \bottomrule
  \end{tabular}}
  \caption{\chg{\textbf{LCA per recording.} LCA@10 and LCA@100 (\%) on learning curves averaged over recordings, as in \cref{tab:etc}, and on the curve of each recording, averaged over the recordings with equal weight per mechanic. \method, from-scratch \method, AdaJEPA, and DreamerV3 average three test-time seeds.}}
  \label{tab:per_recording}
\end{table}

\begin{figure}[t]
  \centering
  \includegraphics[width=\linewidth]{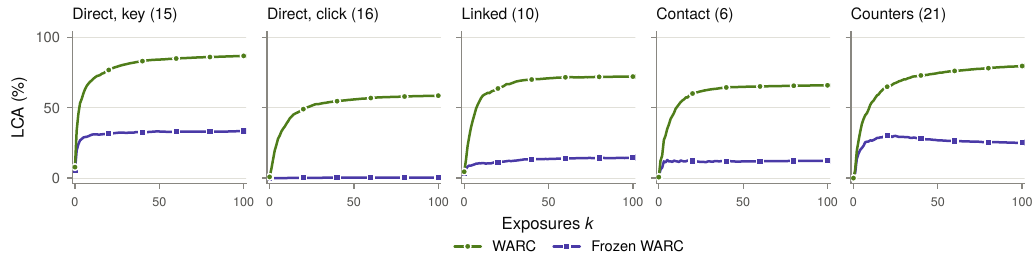}
  \caption{\textbf{Learning curves by kind of mechanic.}
  The mean of LCA$_m$@$k$ over the mechanics of each kind of \cref{tab:mech_kinds} against exposure $k$, up to $k=100$, without the single autonomous mechanic.}
  \label{fig:kinds}
\end{figure}
\paragraph{Kinds of mechanics.}
\Cref{fig:kinds} groups the mechanics by kind.
Frozen \method predicts almost none of the direct click mechanics and every other kind in part.
Test-time training raises this mean for every kind, fastest for the direct key mechanics.

\begin{figure}[p]
  \centering
  \includegraphics[width=\linewidth]{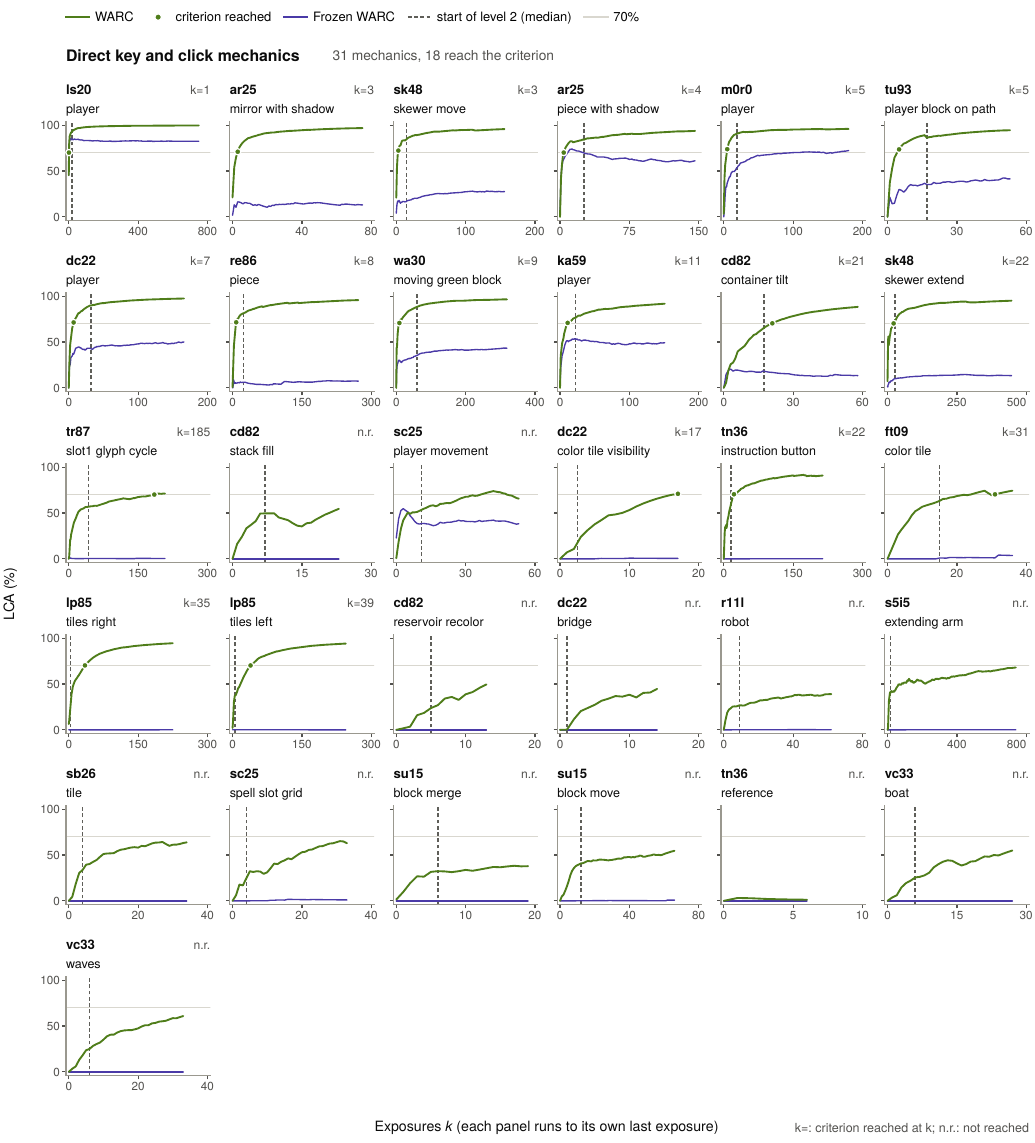}
  \caption{\textbf{The learning curve of every mechanic: direct key and direct click mechanics} (the other kinds in \cref{fig:mech_curves2,fig:mech_curves3}).
  LCA$_m$@$k$ of \method in one test-time seed and of frozen \method against exposure $k$.
  Curves run over every exposure of the mechanic, and a dot marks where \method reaches the criterion.
  Panels are ordered within each kind by the exposure at which \method reaches the criterion, and the mechanics that do not reach it come last.
  The black dashed line marks the mechanic's first exposure in level 2, as the median over the recordings in which it fires in both levels.}
  \label{fig:mech_curves}
\end{figure}
\begin{figure}[p]   %
  \centering
  \includegraphics[width=0.86\linewidth]{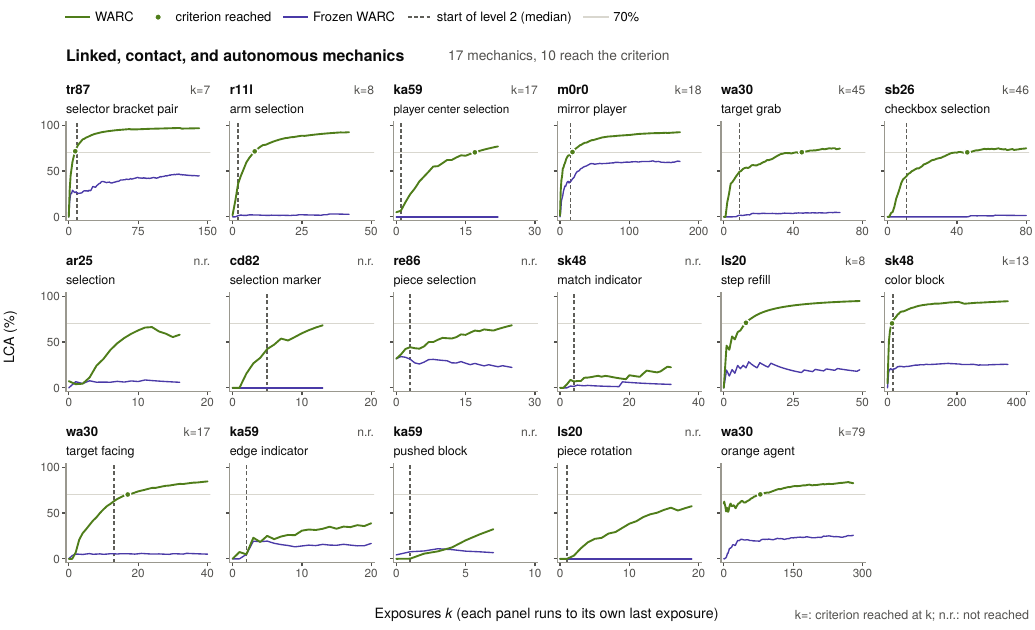}
  \caption{\textbf{The learning curve of every mechanic: linked, contact, and autonomous mechanics}, shown as in \cref{fig:mech_curves}.}
  \label{fig:mech_curves2}
\end{figure}
\begin{figure}[tp]
  \centering
  \includegraphics[width=0.86\linewidth]{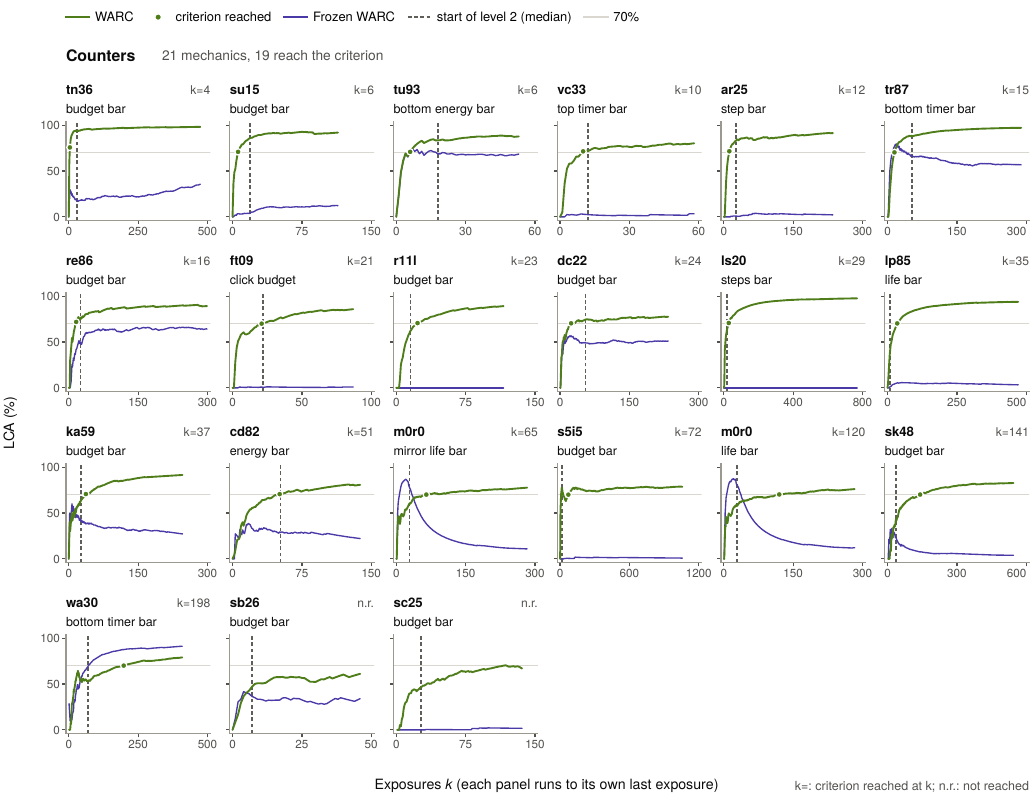}
  \caption{\textbf{The learning curve of every mechanic: counters}, shown as in \cref{fig:mech_curves}.}
  \label{fig:mech_curves3}
\end{figure}

\paragraph{Every mechanic.}
\Cref{fig:mech_curves,fig:mech_curves2,fig:mech_curves3} show the curve of every mechanic for one test-time seed.
\method brings 47, 47, and 48 mechanics to criterion in its three seeds.

\section{Memorization or Generalization}
\label{app:novelty}

This section defines what can be new about an exposure and splits the predictions of \cref{sec:generalization} by it.

\paragraph{Classes.}
Within a recording, each scored exposure with $k\geq1$ gets one class by what is new about it, defined separately for moves, direct click mechanics, and all mechanics.
The history of an exposure is every earlier exposure to the mechanic in the recording, scored or not.
A mechanic's first exposure in level 2 is not classed.
The classes are heuristic, and each covers different mechanics, so classes are compared within a mechanic before averaging, with LCA@100 counted from exposure 1, since exposure 0 is always the first, unless noted.

\paragraph{Moves (direct key).}
The 13 moves are the direct key mechanics in which one of the four direction keys moves, tilts, or extends an object.
The other two direct key mechanics, in cd82 and tr87, change an object in place and are left out of this split.
A move is a \emph{repeat} if the mechanic has fired earlier in the recording under the same key at the same place, that is, with the same start and end positions of the object.
It is a \emph{new position} if the mechanic has fired under the key only at other places, and a \emph{new direction} if it has not fired under the key.

\paragraph{Direct click.}
For the 16 direct click mechanics, the clicked object is the game object under the clicked cell before the click, identified by its name and position.
A click on the background or on an object larger than 1{,}024 cells is identified by the cell itself.
A click is a \emph{repeat} if the same object was clicked at the same cell earlier in the recording, a \emph{new cell} if the same object was clicked at another cell, and a \emph{new object} otherwise.

\paragraph{All mechanics.}
For all 69 mechanics, we compare the pattern of changed cells, taken relative to their bounding box and including the colors.
An exposure is a repeat if the same cells changed in the same way before, a new position if the same pattern appeared at another place, and a new form otherwise.
A new form can be a new effect or the same effect against different surroundings, so this split is coarser than those of the direct mechanics.

\paragraph{Other recordings.}
At each exposure to a mechanic in level 1 of the adapted recording $A$, the model predicts the other recordings before it updates on that exposure.
Each predicted step uses the three frames and actions before it in its own recording.
In level 1, we predict up to 48 steps at which the mechanic fires, taken from the other recordings in turn.
In level 2, we predict every such step, with the model adapted only on level 1 of $A$.
A predicted step is classed against the history of $A$, so a seen transition is a repeat and an unseen one is any other class.
A mechanic is included when it fires in level 1 of $A$ and in the predicted level of another recording, which holds for 14 of the 15 direct key and 14 of the 16 direct click mechanics.
Means give equal weight per mechanic after averaging over the adapted recordings.

\begin{figure}[t]
  \centering
  \includegraphics[width=0.96\linewidth]{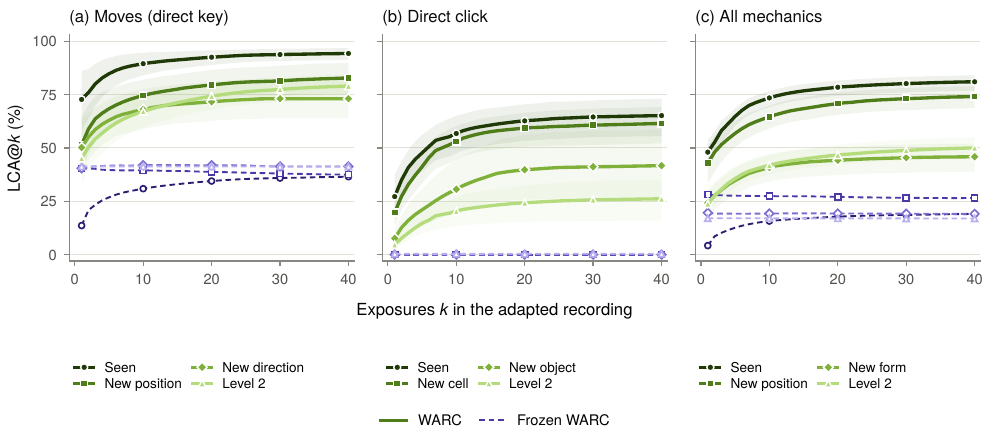}
  \caption{\textbf{Generalization to other recordings, by class.}
  \Cref{fig:generalization} with the unseen transitions split by what is new about them, for (a) the 12 moves among the 14 direct key mechanics of \cref{fig:generalization} and (b) the 14 direct click mechanics that fire in level 1, and (c) all 64 mechanics that fire in other recordings, 63 of them in level 1.
  Bands show 95\% intervals for \method, and dashed lines show frozen \method on the same steps.}
  \label{fig:probes}
\end{figure}

\paragraph{By class.}
In \cref{fig:probes}, at $k=40$, a move from a new position or in a new direction trails a seen move.
A known object clicked at a new cell is predicted about as well as a seen click, whereas a click on a new object trails it.
The paired 95\% intervals of these differences exclude zero except for a click at a new cell.

\begin{figure}[t]
  \centering
  \includegraphics[width=\linewidth]{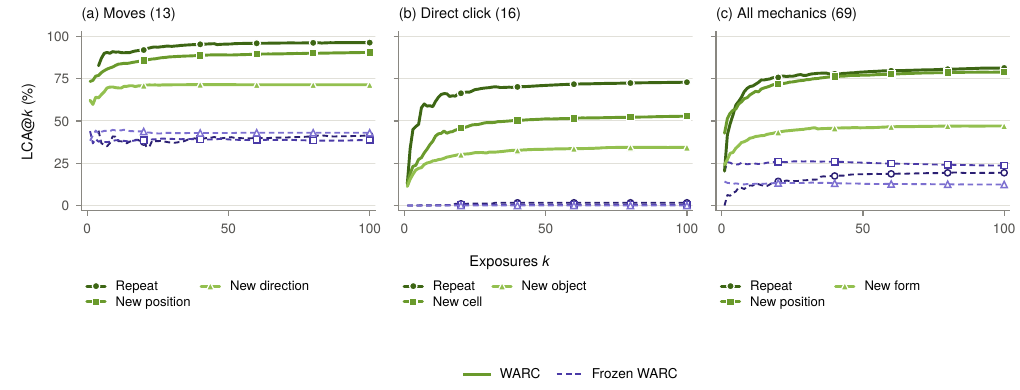}
  \caption{\textbf{Memorization or generalization.}
  LCA@$k$, counted from exposure 1, up to $k=100$, split by what is new about each exposure within its recording: (a) the 13 moves, (b) the 16 direct click mechanics, and (c) all 69 mechanics by the pattern of changed cells.
  Not every class occurs at small $k$.
  Points are drawn only where at least five mechanics have an exposure. \method averages three test-time seeds, and dashed lines show frozen \method.}
  \label{fig:novelty_all}
\end{figure}
\begin{figure}[t]
  \centering
  \includegraphics[width=0.5\linewidth]{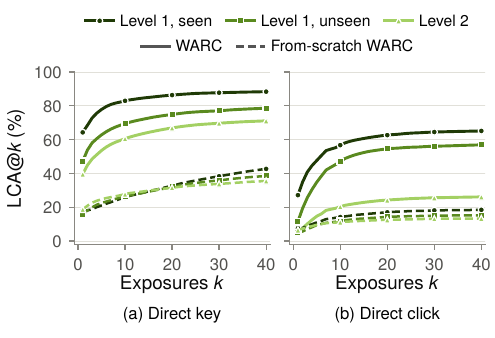}
  \caption{\chg{\textbf{Generalization without pretraining.} \Cref{fig:generalization} with from-scratch \method (dashed) next to \method (solid), on the same probes and averaged over three test-time seeds.}}
  \label{fig:generalization_scratch}
\end{figure}

\paragraph{\chg{From-scratch \method.}}
\chg{\Cref{fig:generalization_scratch} adds from-scratch \method to \cref{fig:generalization}, on the same probes.
Its curves rise more slowly than those of \method in every class and end lower.}

\begin{figure}[t]
  \centering
  \includegraphics[width=\linewidth]{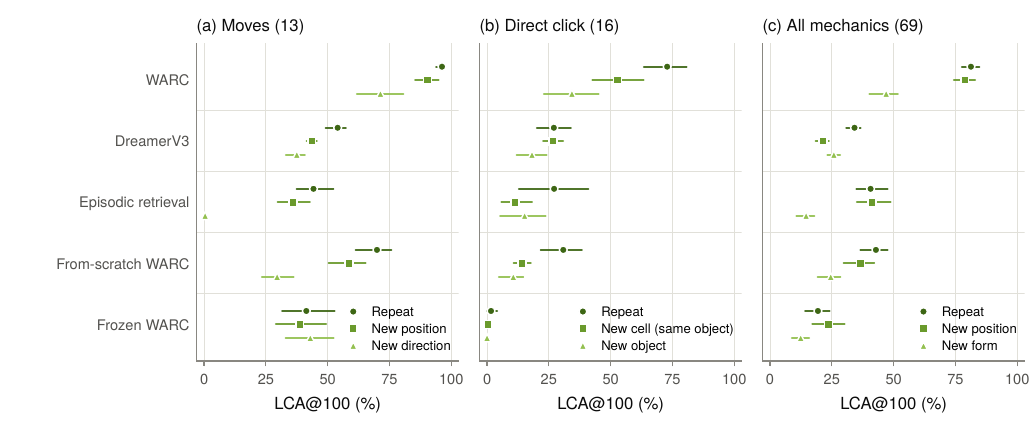}
  \caption{\textbf{The split for other world models.}
  LCA@100, counted from exposure 1, on repeats, new positions or cells, and new directions, objects, or forms, for \method, frozen \method, from-scratch \method, episodic retrieval, and DreamerV3.
  \method, from-scratch \method, and DreamerV3 average three test-time seeds.}
  \label{fig:novelty_models}
\end{figure}

\paragraph{Within a recording.}
\Cref{fig:novelty_all} splits every exposure by its class within its own recording, and \cref{tab:novelty_seeds} gives LCA@100 of each class.
Over all mechanics, a new position is predicted nearly as well as a repeat, and for moves it trails a repeat by a few points.
A new direction, a new cell, a new object, and a new form are predicted much less well than a repeat.
In every seed, the paired intervals of these differences exclude zero except for a new position over all mechanics.
These gaps are measured against a repeat in the same recording, and those of \cref{fig:probes} against a seen transition in another recording, so the two are not directly comparable.
A move in a new direction or from a new position is predicted better than by frozen \method on the same exposures (\cref{fig:novelty_models}a).
The ordering of the classes is the same with LCA@19 or LCA@50, in every seed, and without the 21 counters.

\begin{figure*}[t]
  \centering
  \includegraphics[width=\textwidth]{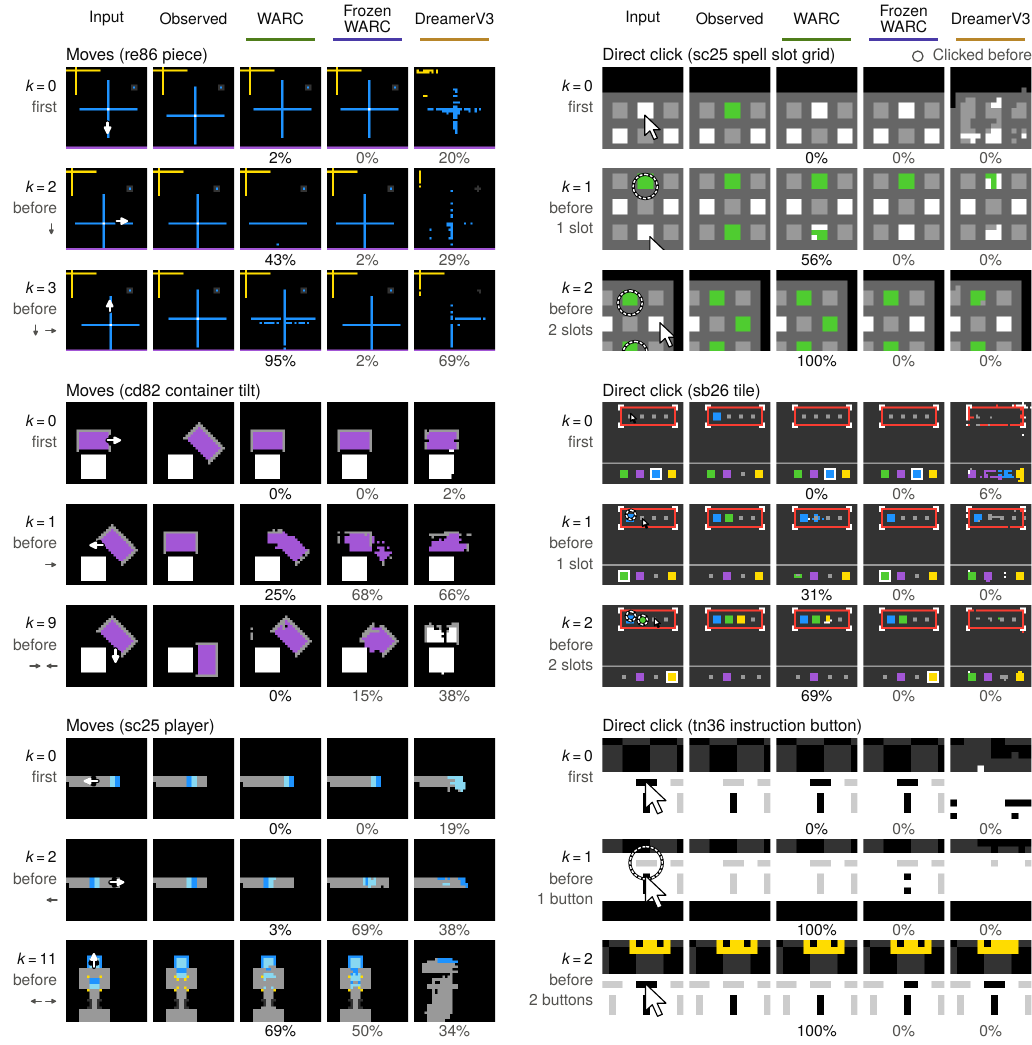}
  \caption{\textbf{More examples of generalization to unseen transitions.}
  \Cref{fig:examples} for three moves (left) and three direct click mechanics (right).
  The last row of sc25 player is in level 2.}
  \label{fig:examples_more}
\end{figure*}
\paragraph{Other models.}
\Cref{fig:novelty_models} applies the same classes to four other models.
From-scratch \method shows the same ordering as \method for moves and direct clicks, and episodic retrieval predicts almost none of the moves in a new direction.

\begin{table}[t]
  \centering
  \footnotesize
  \setlength{\tabcolsep}{5pt}
  \begin{tabular}{@{}llrr@{}}
    \toprule
    Group & Class & LCA@100 & vs.\ repeat \\
    \midrule
    Moves & repeat & 96.2 & -- \\
     & new position & 90.4 & $-5.8$ \\
     & new direction & 71.4 & $-24.9$ \\
    \addlinespace
    Direct click & repeat & 72.9 & -- \\
     & new cell & 52.8 & $-20.1$ \\
     & new object & 34.3 & $-38.5$ \\
    \addlinespace
    All & repeat & 81.3 & -- \\
     & new position & 78.8 & $-2.5$ \\
     & new form & 47.0 & $-34.3$ \\
    \bottomrule
  \end{tabular}
  \caption{The split of \cref{fig:novelty_all}, averaged over three test-time seeds: LCA@100 (\%), counted from exposure 1, with equal weight per mechanic, and the mean difference from repeats in the same seed.}
    \label{tab:novelty_seeds}
\end{table}

\begin{figure}[t]
  \centering
  \includegraphics[width=0.42\linewidth]{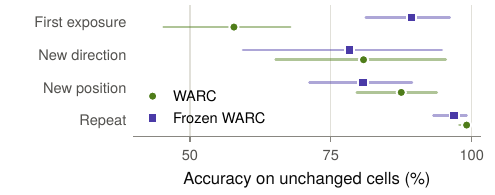}
  \caption{\textbf{Blocked moves.}
  The split of moves in \cref{fig:novelty_all}a for blocked moves (19 types in 12 games), scored on the region's unchanged cells (predicting no change scores 100), at the first blocked move and at five to nineteen exposures, with 95\% intervals.}
    \label{fig:blocked}
\end{figure}
\paragraph{More examples.}
\Cref{fig:examples_more} shows six more mechanics in the format of \cref{fig:examples}, with frozen \method in place of Qwen3.8.
\method does not always predict a move in a new direction, as in cd82 and at the first move to the right in sc25.

\paragraph{Blocked moves.}
Blocked moves are steps at which a move fails because a wall, a boundary, or another object stops it.
The object should stay in place, so a blocked move is scored on the unchanged cells of its region, which covers the blocked object and the position it tried to move to, as the fraction that the model also predicts unchanged.
The false-edit rate of \cref{sec:setup} includes these steps.
Blocked moves are grouped into types, one per kind of obstacle in a game, and classed as moves are, with the earlier blocked moves of a type counted as exposures (\cref{fig:blocked}).
Once \method has learned how an object moves, it predicts the move even the first time a wall or another object blocks it, so the object passes through the obstacle in the prediction.
At these first blocked moves, 58\% of the cells that should stay unchanged are predicted unchanged, against 89\% for frozen \method, and the paired 95\% interval of this difference excludes zero.
At five to nineteen exposures, \method predicts blocked moves at least as well as frozen \method.

\section{Ablation Details}
\label{app:ablations}

\begin{table}[t]
  \centering
  \footnotesize
  \begin{tabular}{@{}llcc@{}}
    \toprule
    Choice & Change & LCA@10 & LCA@100 \\
    \midrule
    Pretraining & none & $-36.8$ [$-39.7$, $-34.0$] & $-37.0$ [$-40.2$, $-34.0$] \\
    Update & every 5 transitions & $-3.6$ [$-5.8$, $-1.5$] & $-1.2$ [$-2.2$, $-0.4$] \\
    Updated part & LoRA, rank 64 & $+2.2$ [$+0.9$, $+3.4$] & $-1.6$ [$-2.7$, $-0.1$] \\
     & last two blocks & $-5.9$ [$-8.0$, $-4.0$] & $-4.3$ [$-5.3$, $-3.5$] \\
     & modulation layers & $-9.6$ [$-12.7$, $-6.7$] & $-9.9$ [$-12.1$, $-7.9$] \\
    Augmentation & none & $-1.5$ [$-3.0$, $0.0$] & $-1.1$ [$-2.1$, $-0.1$] \\
     & + flips and color permutations & $+1.1$ [$-0.1$, $+2.4$] & $+0.2$ [$-0.9$, $+1.4$] \\
     & color permutations instead of rotations & $+0.3$ [$-1.5$, $+2.3$] & $-0.2$ [$-1.6$, $+1.4$] \\
    \bottomrule
  \end{tabular}
  \caption{Paired 95\% bootstrap intervals for no pretraining, the ablations of \cref{tab:ablations}, and three changes to the augmentation. Entries give the change from \method in LCA@10 and LCA@100, averaged over three test-time seeds. Games and then recordings are resampled 2{,}000 times, with the same draws for every seed and row.}
  \label{tab:ablation_ci}
\end{table}
\paragraph{Intervals.}
\Cref{tab:ablation_ci} gives paired intervals for \cref{tab:ablations}.
Every change of \cref{tab:ablations} except LoRA lowers both LCA@10 and LCA@100, with intervals that exclude zero.
LoRA raises LCA@10 and lowers LCA@100 slightly, and both of its intervals exclude zero.

\paragraph{Update cadence and gradient steps.}
Updating every five valid transitions takes 100 gradient steps at each update, so the steps per transition equal those of \method.
In one test-time seed, doubling the gradient steps per transition does not raise LCA@100, and halving them lowers it by 1.0 point.

\paragraph{Updated parameters.}
The rows that restrict the update change which parameters are updated and keep the rest of the recipe of \method, except for LoRA.
The modulation layers are the adaptive layer norms of every block, and the last two blocks include the final norm and the head.
LoRA~\cite{hu2022lora} adds rank-64 adapters, 3.3M parameters, to the attention and MLP linear layers of every block.
AdamW updates the adapters with a learning rate of $10^{-3}$, chosen on the five games outside \evalsuite.
For ranks 8 and 32, each run with one test-time seed, the paired interval of the difference from the full update in LCA@100 includes zero.

\paragraph{Augmentation.}
LCA@100 is lower without augmentation, and the intervals of the other two augmentation rows in \cref{tab:ablation_ci} include zero.
The rotations matter most for moves in a new direction.
Averaged over three test-time seeds, LCA@100 of a move in a new direction trails that of a repeat by 34.7 points without augmentation and by 24.9 with it.
For new positions and for clicks on new objects, the gap changes by less than three points.
We keep rotations, chosen on the five games outside \evalsuite.

\begin{figure}[t]
  \centering
  \includegraphics[width=0.42\linewidth]{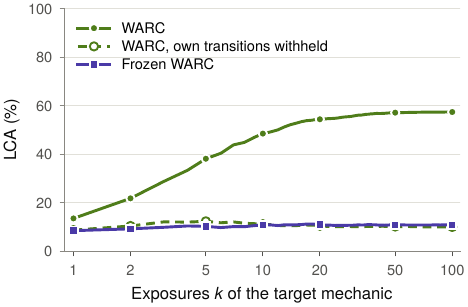}
  \caption{\textbf{Withheld transitions.}
  LCA@$k$ on the 32 target mechanics for \method with their transitions in the buffer or withheld from it, and for frozen \method, against exposure $k$ on a logarithmic axis.}
    \label{fig:controls}
\end{figure}

\paragraph{Withheld transitions.}
In each recording, we choose one target mechanic among those that fire at least five times and in at most a quarter of the transitions, taking the one that has been a target least often in its game so far.
This gives 32 target mechanics in the 144 recordings that have such a mechanic.
The transitions in which the target fires are predicted and scored but never enter the replay buffer, while the number of updates and the input frames are unchanged.
On the targets, \method with the transitions withheld stays close to frozen \method and far below \method with them in the buffer (\cref{fig:controls}), and the paired 95\% interval of its difference from frozen \method in LCA@100 includes zero.

\FloatBarrier
\section{Pretraining Scale}
\label{app:scaling}

\paragraph{Setup.}
This study uses a variant of \method that splits the transformer into a seven-block encoder that does not see the action and three blocks that do, with a learned token per game.
Nine models are pretrained on nested subsets of 20 training games, with the other five public games held out, and each is adapted by test-time training on 1{,}000 transitions of human play in these five games.
One axis keeps all 20 games and adds recordings per game (1, 2, 4, 8, all), and the other keeps all recordings and adds games (2, 4, 8, 16, 20), so the two meet at the full set.
We evaluate on up to 1{,}024 other held-out transitions by exact-frame error, the fraction of next frames with any wrong cell, and by changed-cell error, the fraction of all changed cells that are predicted with a wrong color.

\begin{figure}[t]
  \centering
  \includegraphics[width=0.96\linewidth]{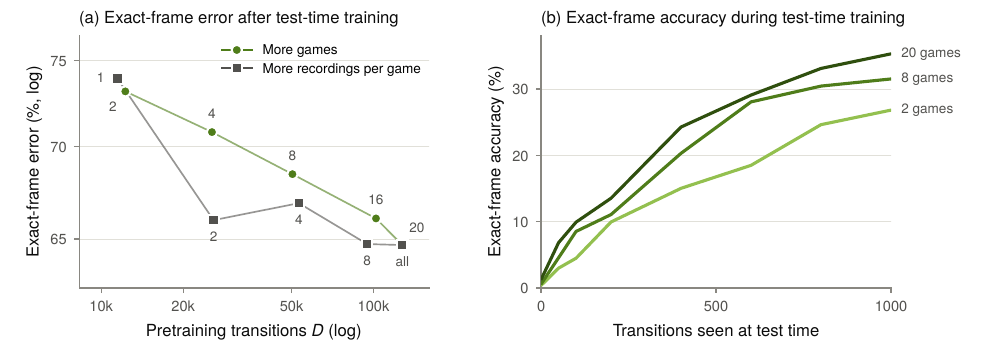}
  \caption{\textbf{Pretraining scale.}
  (a) Exact-frame error, the fraction of next frames predicted with any wrong cell, after test-time training on 1{,}000 transitions, against the number of pretraining transitions $D$.
  Labels give the number of games, or of recordings per game over all 20 games.
  (b) Exact-frame accuracy during test-time training for models pretrained on 2, 8, and 20 games, evaluated at 0, 50, 100, and every 200 transitions.
  Means over five held-out games, one pretraining seed.}
  \label{fig:pretrain_scaling}
\end{figure}

\paragraph{Results.}
Exact-frame error after test-time training falls as pretraining games are added, up to all 20 (\cref{fig:pretrain_scaling}a).
Changed-cell error falls from 35.2\% to 31.5\% with 8 games and then flattens.
At a similar number of transitions, 20 games with a few recordings each give a lower exact-frame error than fewer games with all their recordings, except at the smallest number.
More recordings per game help little beyond two.
Before test-time training, no model predicts more than 3\% of frames exactly, so the gain from broader pretraining appears only after test-time training (\cref{fig:pretrain_scaling}b).
These results come from one pretraining seed and five held-out games, and the test-time training of this study used a recipe that was tuned on these games.

\section{Closed-Loop Details}
\label{app:agent}

\subsection{Agent}

\paragraph{Explorer.}
The explorer is a simplified version of a published graph explorer~\cite{rudakov2025graph}.
It keeps a graph of the states reached in the current level and the actions tried in each.
A state is keyed by a hash of the frame and by the set of available actions.
The hash masks the status bars, thin bars along the top and bottom edges that are located in the first frame of the level.
The actions are the available keys, one click at the center of each connected component of the frame, and a fixed grid of 64 clicks.
They are tried in three priority groups, first the keys and the components of at most 1{,}024 cells, then the grid, then the larger components.
In a state with untried actions, the explorer chooses among the untried actions of the current group, the highest-priority group that still has untried actions.
The choice is uniform when the explorer runs without a world model.
In a state without untried actions, it moves along known edges to the nearest untried action, by priority and then by distance.
If none is reachable, it retries the least-tried actions of the current state.
The published explorer detects status bars by their shape and position and opens the clicks in five groups by color and size.
Our explorer instead orders the clicks by size and masks only thin bars along the top and bottom edges.
It also adds the available actions to the state key and never rebuilds its graph within a level.

\paragraph{Ranking actions with the world model.}
For a frame $o$, let $\phi(o)$ be the color histograms of its $4{\times}4$-cell blocks, with the status bars and the action-independent cells (defined below) masked, and let $d(\phi,\phi')=\tfrac12\lVert\phi-\phi'\rVert_1$.
Let $\mathcal{O}_\ell$ be the distinct frames of the current level and $r$ the median of $d\big(\phi(\hat o_{t+1}),\phi(o_{t+1})\big)$ over the last 20 transitions.
The novelty of a predicted frame $\hat o$ at depth $j$ is
\begin{equation}
  \nu_j(\hat o)=\max\Big(0,\ \min_{o\in\mathcal{O}_\ell} d\big(\phi(\hat o),\phi(o)\big)-j\,r\Big).
  \label{eq:novelty}
\end{equation}
Subtracting $r$ keeps the prediction errors of a still inaccurate model from counting as novelty.
\method predicts the next frame $\hat o_a$ of every untried action $a$ of the current group, or of a random sample of 256 when there are more.
Candidates with the same predicted frame are merged into one.
For the 32 actions with the highest $\nu_1$, \method also predicts a second step, in which the second action $b$ is $a$ itself or one of three other candidates drawn at random.
These actions are scored by
\begin{equation}
  s(a)=\max\Big(\nu_1(\hat o_a),\ \tfrac12\max_{b}\nu_2(\hat o_{a,b})\Big),
  \label{eq:score}
\end{equation}
while every other action keeps $s(a)=\nu_1(\hat o_a)$.
The explorer takes the action with the highest score, or a uniformly random candidate with probability $\epsilon=0.2$.
Until five prediction errors have been recorded, or when every score is zero, the choice is uniform.
$\mathcal{O}_\ell$ and the errors are reset at each new level, and \method updates with its test-time recipe after every valid transition.
Frozen \method runs the same agent without updates, and Graph runs it without a world model.

\paragraph{Action-independent cells.}
A cell is action-independent if every candidate action leads to the same predicted color for it and this color differs from the current frame, as for a timer or a step counter.
The mask extends to the connected components that contain such cells, excluding components of more than 1{,}024 cells and cells on which the predictions disagree.
It is recomputed at every step, skipped when all actions lead to the same predicted frame, and applies only to the novelty score, not to the state keys of the graph.

\subsection{Protocol and Measures}

\paragraph{Protocol.}
Each of the 25 public games is played with one environment seed and 10 agent seeds, and every measure is read from the first 1{,}000 actions of a run, including resets.
Random picks a uniformly random available action.
Random and the from-scratch CNN were run for 4{,}000 actions, of which the first 1{,}000 are used, and the other agents for 1{,}000.
All runs with these seeds are included in the results.
\Cref{tab:closedloop} reports the 20 games of \evalsuite.
The other five games were used to choose the test-time recipe, and they are included only in the results on all 25 games below, \cref{fig:closedloop_l1}, and \cref{tab:explorer_ablation}.

\paragraph{RHAE.}
The official score, Relative Human Action Efficiency~\cite{arcprize2026methodology}, rates each cleared level by the squared ratio of the human players' actions to the agent's, weights later levels more heavily, and normalizes over all levels of a game on a 0--100 scale.
An agent that clears only the first levels, using many more actions than humans, therefore scores below 1.

\subsection{Results}

\paragraph{All 25 public games.}
On all 25 public games, Graph, Graph + frozen \method, and \method-Agent reach an RHAE of 0.149, 0.181, and 0.455.
They clear level 1 in 31.6\%, 34.4\%, and 46.0\% of the runs.
With an oracle world model, the explorer reaches an RHAE of 1.392.

\paragraph{Intervals.}
On the 20 games, when games and seeds are resampled, the gain of \method-Agent over frozen \method has a paired 95\% bootstrap interval of $+0.31$ [$0.09$, $0.57$] in RHAE and of $+14.0$ [$2.0$, $29.0$] points in level-1 clears.

\paragraph{Results per game.}
In five of the 20 games (ar25, ft09, m0r0, s5i5, tu93), \method-Agent clears level 1 in more runs than frozen \method, and in two (ls20, sk48) in fewer.
In the other 13 games the two agree.

\begin{figure}[t]
  \centering
  \includegraphics[width=0.42\linewidth]{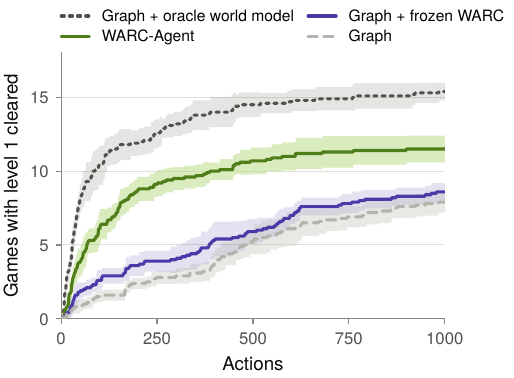}
  \caption{Number of the 25 public games in which level 1 has been cleared within a given number of actions. A pass plays each game once with one seed. Lines: means over the 10 passes. Bands: 95\% intervals of these means, with the games fixed.}
  \label{fig:closedloop_l1}
\end{figure}

\paragraph{Actions to the first clear.}
\Cref{fig:closedloop_l1} shows when each agent first clears level 1.
\method-Agent clears it earlier than frozen \method.
In the runs in which both clear it, \method-Agent needs a median of 60 actions and frozen \method 293.
The difference in games cleared, averaged over budgets up to 1{,}000 actions, has an interval over the 10 passes that excludes zero.
The gap is largest at about 300 actions and narrows later, as the explorer alone also reaches many of these clears.

\begin{table}[t]
  \centering
  \footnotesize
  \begin{tabular}{@{}lccc@{}}
    \toprule
    Explorer & L1@1000$\uparrow$ & L2@1000$\uparrow$ & RHAE@1000$\uparrow$ \\
    \midrule
    Graph + \method (\method-Agent) & 47.2 & 8.8 & 0.456 \\
    \quad without the mask of action-independent cells & 47.2 & 9.6 & 0.513 \\
    \quad a random sample of 32 actions, no mask & 42.4 & 10.4 & 0.346 \\
    \quad one planning step & 44.0 & 8.0 & 0.473 \\
    \quad no random action & 44.8 & 10.4 & 0.520 \\
    \midrule
    Graph + oracle world model & 63.2 & 27.2 & 1.326 \\
    \bottomrule
  \end{tabular}
  \caption{The explorer with components removed, and with an oracle world model, on the 25 public games over five seeds, with the measures of \cref{tab:closedloop}. Every row but the last uses \method with test-time training, and the last row is the oracle world model of \cref{tab:closedloop}.}
  \label{tab:explorer_ablation}
\end{table}
\paragraph{Explorer ablations.}
\Cref{tab:explorer_ablation} removes components of the explorer.
No variant differs from the full explorer beyond its paired interval in level-1 clears or in RHAE, and with 25 games the table rules out only large effects.
We keep the full explorer, since choosing among the variants by their scores in this table would tune the explorer on the evaluated games.

\paragraph{Oracle world model.}
The oracle row of \cref{tab:closedloop} runs the same explorer, actions, and random choices with a perfect world model.
A copy of the game program follows the run, and each candidate action is tried on a further copy, so each predicted frame is the true next frame.
An agent cannot copy its environment, so the oracle row is for analysis only.
In eight of the 20 games, \method-Agent clears no level in any run.
The oracle clears a level in four of them, and in the other four no level is cleared even with the oracle, so there the explorer is the limit.

\paragraph{\chg{\evalsuite on the agent's own play.}}
\chg{\evalsuite can also score the trajectories of \method-Agent.
At each action, we score the frame that \method predicted for the chosen action, before the agent learned from it.
\Cref{tab:agent_mecheval} compares these scores with those on the human recordings of the same games.}

\begin{table}[t]
  \centering
  \footnotesize
  \setlength{\tabcolsep}{5pt}
  \chg{\begin{tabular}{@{}lrrrr@{}}
    \toprule
     & \multicolumn{2}{c}{Agent's play} & \multicolumn{2}{c}{Human recordings} \\
    \cmidrule(lr){2-3}\cmidrule(l){4-5}
    Game & LCA@10 & LCA@100 & LCA@10 & LCA@100 \\
    \midrule
    ar25 & 72.6 & 86.8 & 73.9 & 86.1 \\
    ft09 & 62.1 & 92.2 & 57.6 & 73.9 \\
    ls20 & 83.1 & 85.0 & 56.8 & 70.2 \\
    m0r0 & 36.4 & 83.4 & 62.8 & 80.2 \\
    s5i5 & 48.2 & 80.3 & 52.6 & 59.8 \\
    sk48 & 28.7 & 38.1 & 43.8 & 67.2 \\
    tu93 & 93.1 & 95.4 & 82.6 & 89.2 \\
    \midrule
    All & 54.5 & 74.4 & 58.7 & 74.3 \\
    \bottomrule
  \end{tabular}}
  \caption{\chg{\textbf{\evalsuite on the agent's own play.} LCA@10 and LCA@100 (\%) of \method on the trajectories of \method-Agent and on the human recordings of the same games, over the 20 mechanics that occur in the agent's play. On the human recordings, the LCA is computed on the curve of each recording and averaged over the recordings. The agent's runs use seed 0 and end after 378 to 599 actions.}}
  \label{tab:agent_mecheval}
\end{table}

\FloatBarrier
\section{Additional Related Work}
\label{app:related}

\paragraph{Video world models.}
Video world models trained on large video and interaction datasets simulate the physical world~\cite{yang2024unisim,nvidia2025cosmos} and games~\cite{bruce2024genie,valevski2024gamengen,alonso2024diamond}.
Latent world models such as DINO-WM~\cite{zhou2025dinowm} and V-JEPA 2~\cite{assran2025vjepa2} predict in the feature space of a frozen pretrained encoder.
These models are trained on their target domain, finetuned for a new one, or applied to it without updates.
We instead study how a world model adapts online to an unseen game.

\paragraph{Test-time training and continual learning.}
Dynamic evaluation adapts a sequence model by gradient steps to the sequence it is predicting~\cite{krause2018dynamic}, as \method does on the stream of transitions (\cref{sec:ttt}).
Over a stream of updates, continual learning limits forgetting by replaying earlier data~\cite{rolnick2019replay}, and \method replays earlier transitions in every update.

\paragraph{Exploration.}
Plan2Explore plans toward states where an ensemble of one-step predictors disagrees~\cite{sekar2020plan2explore}, and Go-Explore keeps an archive of visited states and returns to promising ones before exploring from them~\cite{ecoffet2021goexplore}.
Our agent combines a graph of visited states with a ranking of untried actions by the novelty of their predicted outcomes.

\paragraph{ARC puzzles and agents on ARC-AGI-3.}
On the static ARC puzzles, RoSeViT~\cite{liu2026reasoningmodality} and TraceViT~\cite{liu2026tracevit} keep the canvas of VARC, and Loop-OWM learns the rule of a puzzle as an object-centric world model~\cite{gao2026loopowm}.
On ARC-AGI-3, besides the agents cited in \cref{sec:related}, OPINE-World~\cite{courtis2026opine}, Tycho~\cite{lehmann2026tycho}, and DreamTeam~\cite{sarafian2026dreamteam} have a frontier language model write and repair a programmatic world model, and \citet{rodionov2026ablation} ablates the executable model and its verification.
Others give the language model a programmatic or long-term memory~\cite{fox2026prolong,furgale2026nooa}.
Agents without a language model explore with a graph over states~\cite{rudakov2025graph} or train a predictor from scratch on each level, like StochasticGoose~\cite{smit2025goose,arcprize2025preview}.
We compare with the last two in \cref{sec:closedloop}.

\paragraph{Rule changes and rule discovery.}
NovGrid~\cite{balloch2022novgrid}, NovelGridworlds~\cite{goel2021novelgridworlds}, NovelGym~\cite{novelgym2024}, and NovPhy~\cite{gamage2023novphy} inject controlled novelties, such as new objects or changed dynamics, into grid worlds and physics puzzles.
DiG-bench makes rule discovery the task itself~\cite{battleday2026digbench}.
\evalsuite instead follows each mechanic of an existing game across its exposures.

\end{document}